\documentclass[sigconf]{acmart}
\usepackage{booktabs} % For formal tables
\usepackage{tabularx}
\usepackage{multirow}
\usepackage{soul}
\usepackage{amsmath}
\usepackage{amssymb}
\usepackage{enumitem}
\usepackage{mathtools}

\setcopyright{rightsretained}
\acmDOI{10.475/123_4}

\acmISBN{123-4567-24-567/08/06}

\acmArticle{4}
\acmPrice{15.00}

\begin{document}
\title{Adversarial Adaptation of Scene Graph Models for
Understanding Civic Issues}
% \titlenote{Produces the permission block, and
% \subtitle{Extended Abstract}
% \subtitlenote{The full version of the author's guide is available as
%   \texttt{acmart.pdf} document}

\author{Shanu Kumar}
\orcid{}
\affiliation{%
  \institution{Indian Institute of Technology Kanpur, India}
}
\email{sshanu@iitk.ac.in}

\author{Shubham Atreja, Anjali Singh, Mohit Jain}
\affiliation{%
  \institution{IBM Research AI, India}
}
\email{{satreja1,ansingh8,mohitjain}@in.ibm.com}

% \authornote{This author is the
%   one who did all the really hard work.}
% \affiliation{%

% \affiliation{%
% \affiliation{%
% \affiliation{%
%   \country{China}
% }

% \affiliation{%


% \author{Julius P.~Kumquat}


\begin{abstract}
Citizen engagement and technology usage are two emerging trends driven by smart city initiatives.
Governments around the world are adopting technology for faster resolution of civic issues.
Typically, citizens report issues, such as broken roads, garbage dumps, \textit{etc.} through web portals and mobile apps, in order for the government authorities to take appropriate actions.
Several mediums -- text, image, audio, video -- are used to report these issues.
Through a user study with 13 citizens and 3 authorities, we found that image is the most preferred medium to report civic issues.
However, analyzing civic issue related images is challenging for the authorities as it requires manual effort.
Moreover, previous works have been limited to identifying a specific set of issues from images.
In this work, given an image, we propose to generate a \textit{Civic Issue Graph} consisting of a set of objects and the semantic relations between them, which are representative of the underlying civic issue. We also release two multi-modal (text and images) datasets, that can help in further analysis of civic issues from images. We present a novel approach for adversarial training of existing scene graph models that enables the use of scene graphs for new applications in the absence of any labelled training data. We conduct several experiments to analyze the efficacy of our approach, and using human evaluation, we establish the appropriateness of our model at representing different civic issues.
\end{abstract}




\maketitle

\section{Introduction}
\label{introduction}

In recent years, there has been a significant increase in smart city initiatives \cite{cocchia2014smart,nam2011conceptualizing,neirotti2014current}.
As a result, government authorities are emphasizing the use of technology and increased citizen participation for better maintenance of urban areas.
% Governments are actively working towards creating channels for the citizens to voice their opinion. 
% For instance, Mobile Crowd Sensing (MCS) is used to gather data on various civic issues \cite{guo2014participatory,yadav2013human}.
Various web platforms -- SeeClickFix~\cite{seeclickfix}, FixMyStreet \cite{fixmystreet}, ichangemycity \cite{icmc-link} -- have been introduced across the world, which enable the citizens to report civic issues such as poor road condition, garbage dumps, missing traffic signs, \textit{etc}., and track the status of their complaints.
Such initiatives have resulted in exponential increase in the number of civic issues being reported~\cite{mayor-report}. 
Even social media sites (Twitter, Facebook) have been increasingly utilized to report civic issues.
Studies have found the importance of civic issue reporting platforms and social media sites in enhancing civic awareness among citizens \cite{skoric2016social}.
These platforms help the concerned authorities to not only identify the problems, but also access the severity of the problems.
Civic issues are reported online through various mediums -- textual descriptions, images, videos, or a combination of them.
Previous work \cite{dahlgren2011-modalities} highlights the importance of mediums in citizen participation.
Yet, no prior work has tried to understand the role of these mediums in reporting of civic issues.
% One of the reasons behind the success of these platforms is the high penetration of Internet and smartphones~\cite{311}. In New York, the number of issues reported through mobile and web interfaces have increased significantly, from 19.6\% in 2014 to 47.5\% in 2017 \cite{mayor-report}.
% There are various platforms to report these issues, such as social media sites (Twitter, Facebook), and complaint forums (FixMyStreet\footnote{http://fixmystreet.com/}, ichangemycity\footnote{https://www.ichangemycity.com/}), 
% % http://swachh.city/}, 
% wherein issues can be reported through textual descriptions, images, videos, or a combination of them.

In this work, we first identify the most preferred medium for reporting civic issues, by conducting a user study with 13 citizens and 3 government authorities.
Using the 84 civic issues reported by the citizens using our mobile app, and follow-up semi-structured interviews, we found that images are the most usable medium for the citizens.
In contrast, authorities found text as the most preferred medium, as images are hard to analyze at scale.

% In some cases, citizens may be required to provide additional details such as location and issue category, while in other cases, the authorities determine such information manually.
% Some tasks can be automated through existing machine learning techniques, such as NER for location detection, supervised classification for determining issue category etc \cite{citicafe} 

%As stated earlier, the information can be present in two different media (images and text) and before applying any of the existing machine learning techniques, it is important to understand the representation and purpose of each media and how they interact with each other. Particularly, this has to be understood from the point of view of the stakeholders, ie the citizens who are reporting the issues and the authorities who are acting upon this issue, so that the appropriate technology can be identified to perform the tasks. 
%The primary objective of this work is to understand this dynamics of how the information is collected by the citizens and processed by the authorities and based on our understanding, propose a new way to process this information for detecting civic issues. 
%First, we conduct a user study to understand i) how citizens utilise various media to provide information about issues and ii) how the authorities try to process this information. From the user study, we infer that images play an important role in the representation of a civic issue.

\begin{figure}
  \includegraphics[width=0.47\textwidth, height = 2.3in]{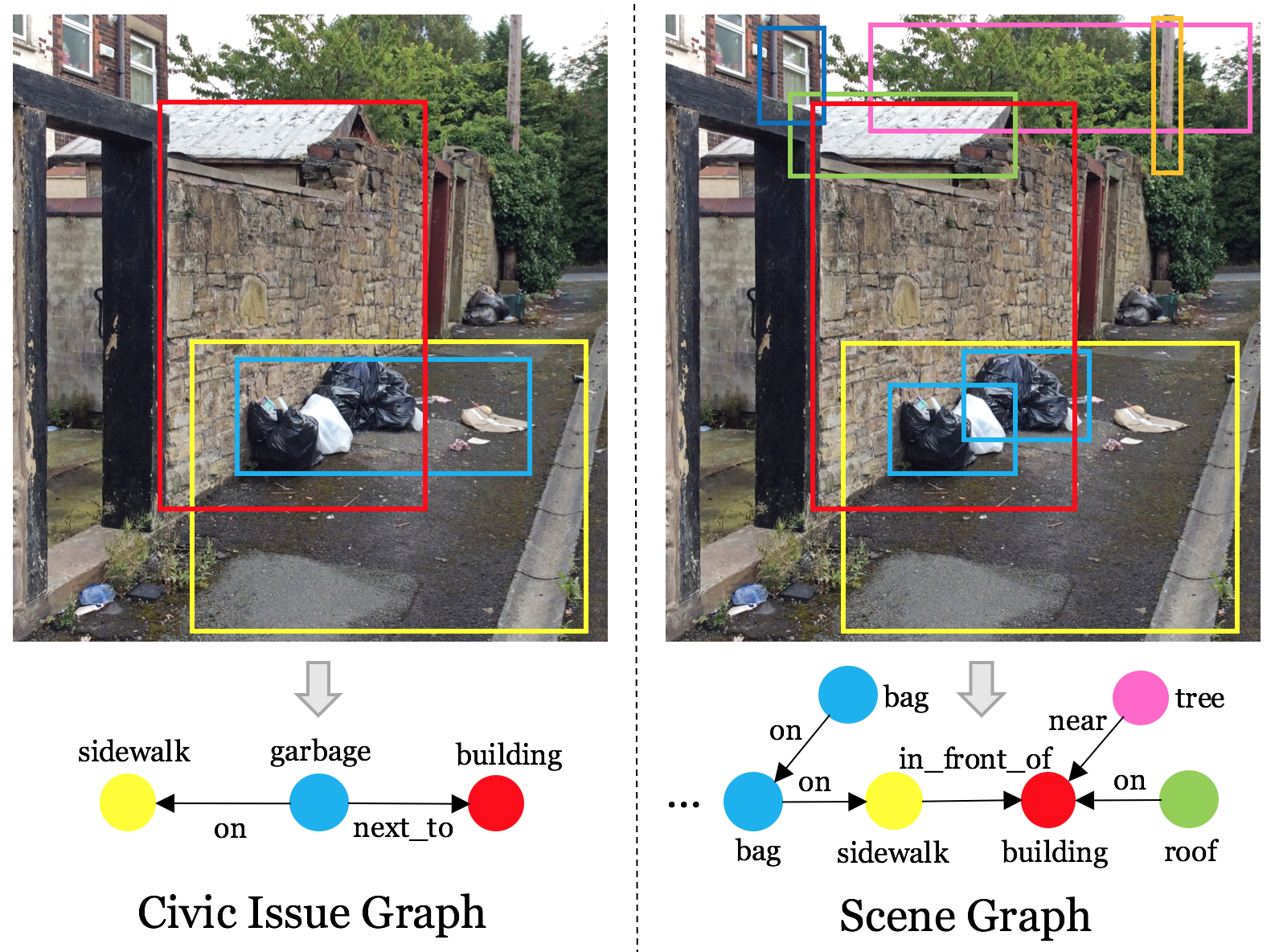}
 \caption{Comparison between Civic Issue Graph and Scene Graph for the same image. The scene graph provides a complete representation of all objects and relationships in the image, while the Civic Issue Graph only consists of relations representative of the civic issue.  }
  \label{fig:intro-pic}
\end{figure}

% \todomj{this para needs more work. i am not sure what it is trying to convey.}
% To fill this gap, there are several ways to analyze images and detect civic issues from them using image understanding techniques. Previous work 
To fill this gap, several works have proposed methods to automatically identify a specific category of civic issues from images, such as garbage dumps \cite{mittal2016spotgarbage} and road damage \cite{road1}.
However, their methods are limited to the specific categories that they address.
Furthermore, existing holistic approaches of analyzing civic issues are limited to text \cite{citicafe}.
To this end, we propose an approach to understand various civic issues from input images, independent of the type of issue being reported.
% However, previous work \cite{dumrewal2018citicafe} highlights inherent problems with categorization of civic issues, such as \todomj{mention the problems}.
% Through our experiments, we further highlight this confusion by implementing one such classification technique.
% Another way to understand images for civic issue detection is through object detection \cite{mittal2016spotgarbage,deep2}.
% However, such techniques cannot be extended to all types of civic issues. For example, consider the following issues: `\textit{tree growing over the fence}', `\textit{water logging around a manhole}', `\textit{garbage dumped on sidewalk next to a house}'.
% To understand these issues properly, it is imperative to identify the right set of objects as well as the semantic relations between them. 

% There can be multiple ways to analyze images and detect civic issues from them, the simplest being -- using supervised classification based on visual features and tagging each image with a civic issue category. Such a classification requires a set of well defined categories along with some training examples for each category. Some works \cite{mittal2016spotgarbage,deep2} propose object detection techniques to identify objects in an image which are indicative of specific civic issues like garbage and road-cracks.

% Scene Graphs are structured representations of an image consisting of visually grounded objects and their relationships from the image \cite{messagepassing}. 

One of the latest advancements in the field of image understanding is generation of scene graphs \cite{johnson2015image}, with the objective of getting a complete structured representation of all objects in an image along with the relations between them.
However, to understand a civic issue, only certain crucial objects need to be detected, along with the relations between them, which are representative of the civic issue in the image.
Inspired from the task of scene graph generation, we propose to generate \textit{Civic Issue Graphs} that provide complete representations of civic issues in images.

Figure \ref{fig:intro-pic} shows a comparison between the two representations. In contrast to the scene graph, the Civic Issue Graph only consists of objects conveying a civic issue, their bounding boxes, and the predicate between these objects. We present a formal definition of this representation in Section \ref{cigg}. 

Training a scene graph model requires a large amount of data consisting of images with grounded annotations (objects and relations in the images). Due to the lack of sufficient annotated images of civic issues, we use an existing scene graph model in a cross-domain setting, with partially annotated and unpaired data.
We utilize a dataset extracted by collating and processing public datasets of images from civic issue complaints, for training our model, and make this dataset publicly available\footnote{Link hidden for blind review}. We present a novel adversarial approach that uses an existing scene graph model for a new task in the absence of any labelled training data. 
%Finally, through predicate fine-tuning, we further refine our model, to generate the most appropriate set of relations that depict the issue in a given image.
% (explain in Section \ref{sec:dataset}) 
To the best of our knowledge, this is the first attempt at adversarial training of an existing scene graph model. We conduct various experiments to establish the efficacy of our approach using metrics derived from standard scene graph evaluation metrics. Finally, through human evaluation, we demonstrate that civic issues from images can be appropriately represented using our Civic Issue Graph.

To summarize, the major contributions of this paper are:
(i) understanding the usability of different mediums for reporting civic issues,
(ii) introducing a novel unsupervised mechanism using adversarial adaptation of existing scene graph models to a new domain,
(iii) experimental evaluation which shows significant performance gains for identification of civic issues from user uploaded images, and
(iv) releasing two multi-modal (text and image) datasets with information on civic issues, to encourage future work in this domain.
%%Existing works for scene graph generation require a large set of images with grounded annotations for objects and the relations but getting this data for a domain specific setting is expensive. We use a pre-trained scene graph model and adapt it to our task in a low cost setting (eg. using partially annotated, loosely paired data). We release two datasets that can 
%We propose to use Adversarial Discriminative Domain Adaption (ADDA) to adapt existing scene graph model for our new task. 

% PUT IN RELATED WORK: Some works \cite{zhuang2018hcvrd,liang2018visual} utilize zero shot learning approaches for identifying unseen relations while constructing a scene graph, but they do not provide a way of, but their experiments show that the learning is restricted to the task of classifying new predicates which were not seen during the training phase

%For the adversarial training approach to work, we still require some data for the model to understand the new representation (CG). We explain the process of extracting this information from public datasets about complaints in Section \ref{sec:dataset}. 

% We now summarize the major contributions of our paper:
% i) Understanding the usability of different mediums to report civic issues through a user study,
% ii) Analyzing images of civic issues using a novel method of adapting existing scene graph models to a new domain, and 
% iii) Two image image datasets with information specific to civic issues, to support future work in this domain.

\section{Related Work}
\label{relatedWork}
% we divide related work into 3 parts: work specific to civic domain, CS work for civic domain, work on scene graphs
\textbf{Civic Issue Detection and Analysis.} 
% Recently, several efforts have been made to collect and analyze data corresponding to the wide range of civic issues that exist in urban landscapes, such as garbage dumps \cite{mittal2016spotgarbage}, road damage \cite{road1,road2}, etc. 
Traditionally, different methods have been employed that use technology to gather data about civic issues: such as using laser imaging to identify uneven roads \cite{pothole1} or gathering data from GPS sensors \cite{pothole2} for detecting potholes. However, these methods are specific to a particular type of civic issue and require a technological setup with additional costs, which may not be convenient at a larger scale. More recently, social media has provided a convenient interface that allows citizens to report civic issues \cite{agostino,karakiza}. Several works try to analyze online platforms to automatically mine issues related to civic amenities \cite{socialmedia1,socialmedia2}, but the analysis is limited to textual descriptions. Specific to images, \cite{road1} and \cite{mittal2016spotgarbage} use object detection and image segmentation techniques to identify road damage and garbage dumps respectively, from input images. However, their methods are also limited to the specific category of civic issues that they address. One of the more recent works, `Citicafe' \cite{citicafe} goes a step ahead by allowing users to report various types of civic issues and further employs machine learning techniques to understand and analyze the civic issue from the user input.
However, they do not provide a method for understanding images reporting different types of civic issues, as we do in this paper.
\\
% With the aim of covering as many types of civic issues as possible, we also use images as the basis for identifying civic problems.

\noindent \textbf{Scene-Graph Generation.} Several works \cite{neuralmotif,klawonn2018generating} propose methods for generating scene graphs from images to represent all objects in an image and the relationships between them. One approach \cite{sg1} includes aligning object, phrase, and caption regions with a dynamic graph based on their spatial and semantic connections. Another approach \cite{messagepassing} uses standard RNNs and learns to iteratively improve its predictions via message passing. Zellers et al. \cite{neuralmotif} present a state-of-the-art technique by first establishing that several structural patterns exist in scene-graphs (which they call motifs) and showing how object labels are highly predictive of relation labels by analyzing the Visual Genome dataset \cite{visualgenome}.
All of these approaches require a large set of images for training with grounded annotations for objects and relations. Some works \cite{zhuang2018hcvrd,liang2018visual} utilize zero shot learning for generating a scene-graph. However, their results show that the learning is restricted to the task of detecting new predicates which were not seen during the training phase. Our approach can be used to generalize existing scene graph models to predict new relations belonging to a different domain, which are absent from the training data.
\\
%such as aligning object, phrase, and caption regions with a dynamic graph based on their spatial and semantic connections \cite{sg1}, using standard RNNs and learning to iteratively improve its predictions via message passing \cite{messagepassing}, etc. Zellers et al. first establish that several structural patterns exist in scene-graphs, which they call motifs and secondly show how object labels are highly predictive of relation labels by analyzing the Visual Genome dataset \cite{visualgenome}.

% \noindent \textbf{Domain Adaptation.} Domain adaptation is a long studied problem, where approaches range from fine-tuning networks with target data \cite{da1} to adversarial domain adaptation methods \cite{da6}. Adversarial methods take a different approach by using a domain classifier to learn mappings from the source domain to target domain and utilize it to generalize the model to the target domain. Adversarial methods have shown promising results for other image understanding tasks such as captioning \cite{da7} and we propose to use a similar method for adapting scene graph models to our new task. 
% ]

%Domain adaptation is a long studied problem where a classifier is learned using labeled data in source domain for unseen or unlabeled data in the target domain, in the presence of a shift in the data distribution of both domains.

\noindent \textbf{Domain Adaptation.}  Domain adaptation is a long studied problem, where approaches range from fine-tuning networks with target data \cite{da1} to adversarial domain adaptation methods \cite{da6}. Some of the deep learning methods propose to learn a latent space that minimizes distance metrics such as maximum mean discrepancy (MMD) \cite{da3} between the source and target domains. A different approach involves domain separation networks which learn to extract image representations that are partitioned into two subspaces: one component which is private to each domain and one which is shared across the two domains \cite{da4}.
%Recent methods \cite{da3}, \cite{da6}, \cite{da8} perform unsupervised domain adaptation by learning source and target feature representations, such that the final classification predictions are made based on features that are both discriminative and invariant to the change of domains, i.e. features have similar distributions in the source and target domains. 
%Several methods \cite{da10}, \cite{da11}, \cite{da12} have reduced Maximum Mean Discrepancy (MMD) \cite{da9} loss which computes mismatch between the two data distributions for domain adaptation.

The more recently introduced adversarial domain adaptation methods \cite{da6, da8, da13} take a different approach by using a domain classifier to learn mappings from the source domain to target domain, which are used to generalize the model to the target domain.
% Adversarial Discriminative Domain Adaptation (ADDA) \cite{da6} method learns source and target mappings by inverting domain labels in domain classification loss which confuses the domain classifier.
Adversarial methods have shown promising results for image understanding tasks such as captioning \cite{da7} and object detection \cite{da15}.
Hence, in this work, we propose to use Adversarial Discriminative Domain Adaptation (ADDA) \cite{da6} for adapting scene graph models to our new task.

\section{User Study}
\label{userStudy}

We conducted a user study to understand the preference of different mediums -- text, audio, image and video -- to report civic issues, both from citizens and authorities perspective.
For this, we developed a custom Android app, with the landing page having four buttons, each corresponding to the four mediums.
% , followed by the `Submit' button.
To report an issue, any of the medium(s) could be used any number of times, \textit{e.g.,} a report can comprise of 1 video, few lines of text, and 2 images.
% Only after all the data is gathered, the user needs to click the Submit button.

13 participants (9 male, 4 female, age=28.5$\pm$6.1 years) reported civic issues over a period of 7-10 days.
All the participants were recruited using word-of-mouth and snowball sampling.
All of them were experienced smartphone users, using it for the past 6.2$\pm$2.2 years, and well educated (highest education: 1 high school, 3 Bachelors, 6 Masters, 4 PhDs).
However, only two of them have previously reported civic issues on online web portals.
At the end of the study, a 30-mins semi-structured interview was conducted, to delve deeper into the reasons for (not) using specific medium(s).
Participants were also asked to rate each of the mediums they used on a 5-point Likert scale from NASA-TLX questionnaire~\cite{nasa-tlx} along with providing subjective feedback.
% or combination of mediums.
Participants were not compensated for participation.

Furthermore, we interacted with 3 government authorities (3 male, age = 35-45 years) for 30-mins each, to understand their perspective on the medium of the received complaints.
All interviews were audio-recorded, and later transcribed for analysis.

\subsection{Results}
Overall, 84 (6$\pm$3.7) civic issues were reported by the 13 participants, mainly in the category of garbage (11/13 participants), potholes causing water-logging (9), blocked sidewalk (6), traffic (5), illegal car parking (3), and stray dogs (3).
81 of these issues consisted of image, text, or their combination, while only 2 had audio and 1 had video. Hence, here we only focus on image and text as preferred mediums.

A majority of the participants (10/13) found image to be the best medium for reporting civic issues, followed by text (2/13) and video (1/13).
Images were preferred mainly because it is quick and easy to click an image, and they convey a lot of information:
``\textit{An image is worth 1000 words.}''-P\textsubscript{4}, 
``\textit{its super quick to take pics... even when I pause at a traffic signal, I can take a pic}''-P\textsubscript{10}.
Participants also felt that images are best for conveying the severity of a civic problem. 
They took multiple images from different angles to show the severity of various issues, such as amount of garbage, size of potholes, \textit{etc}.
Interestingly, participants thought that images can ``\textit{act as a proof of the problem... as images don't lie}''-P\textsubscript{6}.
On the other hand, participants complained that people might `\textit{bluff}'/`\textit{exaggerate}' when reporting issues using text.

However, participants complained that images can not be used to capture the temporal variations of civic problem, \textit{e.g.}, ``\textit{images can't say that this garbage has been here for the past week}-P\textsubscript{2}.
% and clicking images may have associated privacy/security issues especially at airports or government buildings.
% Participants also felt burdened to report issues in `\textit{real-time}' which is hard when they are engaged in other activity such as driving.
For this, participants favored text medium, as it enables providing details about the temporal variations of an issue.
% , and ``\textit{can even report an issue later by writing}''-P\textsubscript{2}.
But participants also found that texting requires more time and effort, compared to clicking images.

When participants were asked to choose the best combination of mediums for reporting civic issues, majority of them (9/14) chose image with text.
The combination allows them the freedom to show severity and truthfulness of the issue using image, along with adding other details in text.
Interestingly when participants were asked to think from the perspective of a government authority, a majority of them (6/13) found text to be the best medium, followed by image (4/13).
The main reasons identified by participants were ``\textit{with huge amounts of data, text is much easier to analyze}''-P\textsubscript{7} and at times, images may not be self-explanatory.

\begin{figure}[]
\centering
  \includegraphics[width=0.9\columnwidth]{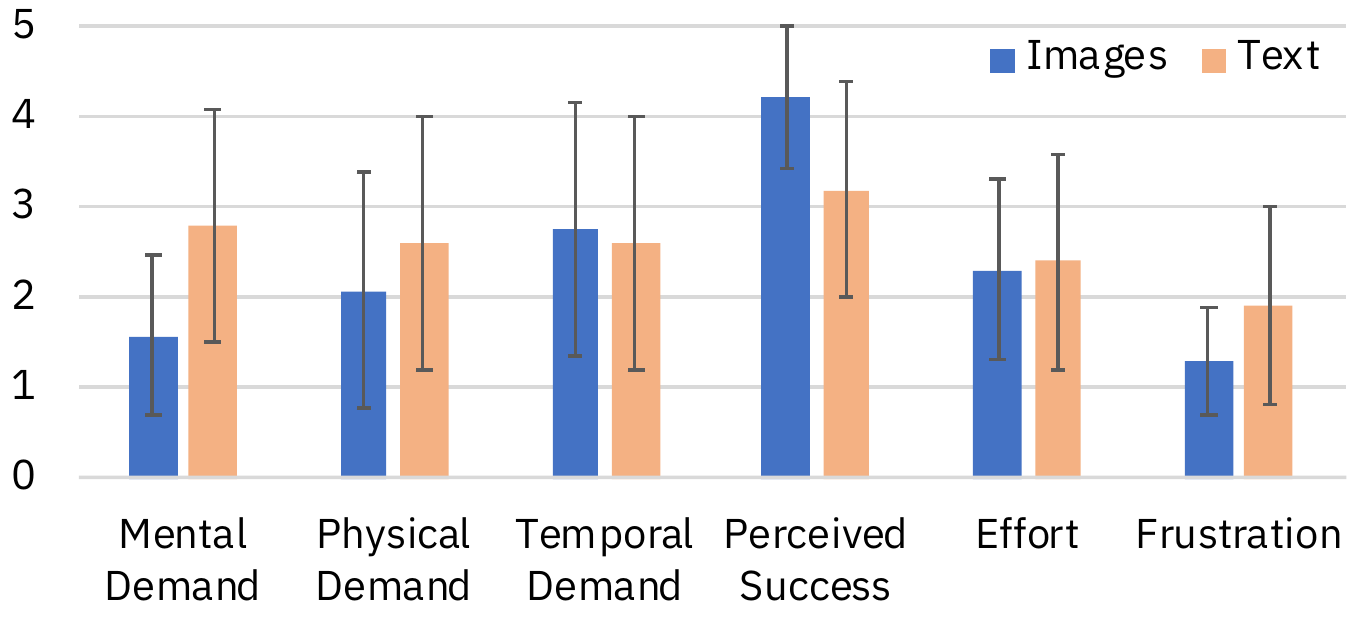}
  \vspace{-2mm}
  \caption{NASA-TLX Likert-scale ratings for Image and Text.}
  \label{fig:rating}
  \vspace{-4mm}
\end{figure}

Participants' responses to the 5-point NASA-TLX Likert scale questions for images and text are shown in Figure \ref{fig:rating}, with the error bars showing the standard deviations.
For all metrics, except perceived success, lower score is better.
As only a few participants used audio (2/13) and video (3/13), we do not discuss their ratings.
A paired t-test showed that images were reported to be significantly better than text, with respect to mental demand (t\textsubscript{12}=3.56, p<0.005) and perceived success (t\textsubscript{12}=2.7, p<0.01).
Only in temporal demand, images performed poorly compared to text, though the difference was not statistically significant.
This was because at times participants had to rush/hurry to click the right image.

Following this, we interviewed 3 government authorities, and found about the process of human annotators analyzing the received civic issue image to generate tags and captions describing the issue. These complaint tags are then passed on to the relevant authority in writing or via phone calls to take appropriate actions.
Also the authorities confirm that a majority of the received complaints comprise of images. However, these images never reach them due to lack adequate technological infrastructure. 
% however they are ``\textit{ambiguous}'', which requires the authority to call the person who raised the complaint in order to get more details.
This confirms that image is the most preferred medium for users, but authorities rely only on textual complaints.
To bridge this gap, in this work, we generate text-based descriptions of images that are used for reporting civic issues.

\section{Dataset}
\label{sec:dataset}

\begin{table}[ht!]
\centering
\begin{tabular}{@{} c *2c @{}}
\toprule
\multicolumn{1}{c}{\textbf{Object Class}} &  \textbf{\#Images} & \textbf{\#Bounding boxes}\\
\midrule\midrule
Garbage & 650 & 831  \\
Manhole & 374 & 419 \\
Pothole & 518 & 677\\
Water logging & 290 & 375 \\
\textbf{Total} & \textbf{1505}  & \textbf{2302} \\
\bottomrule
\end{tabular}
\caption{Statistics for Dataset-1}
\label{table:dataset}
\end{table}

An extensive dataset of images with annotations for a wide variety of civic issues is currently unavailable. To this end, we mined 485,927 complaints (with 131,020 images) from two civic issue reporting forums -- FixMyStreet \cite{fixmystreet} and ichangemycity \cite{icmc-link}. We use them to generate two datasets.

Dataset-1 consists of human-annotated images with the bounding boxes and object labels for 4 object categories (Table \ref{table:dataset}) belonging to the civic issue domain. Some of these object categories are not present in any publicly available image datasets. We utilize the annotations from two existing datasets for garbage \cite{mittal2016spotgarbage} and potholes \cite{road1}, and add new images representative of the new object categories along with their annotations, to build Dataset-1.

Dataset-2 consists of examples of Civic Issue Graphs, represented through triples of the form $[object1, predicate, object2]$, specifying the relationship ($predicate$) between a pair of objects ($object1$ and $object2$). We use natural language processing techniques \cite{corenlp,schuster2015generating} to extract these triples from complaint descriptions. We manually define a set of 19 target object categories which are relevant to the civic domain and map the objects from these triples to our set of target objects using semantic similarity\footnote{http://swoogle.umbc.edu/SimService/GetSimilarity}. We retain only those triples where the predicate defines positional relations (manually determined) and for which both objects are matched with a similarity value greater than 0.4. This dataset consists of 44,353 Civic Issue Graphs, where 8204 are paired with images. There are total 5799 unique relations with 19 object classes and 183 predicate classes.

\section{Civic Issue Graph Generation}
\label{cigg}
We now present our approach for understanding civic issues from input images. We first present the formal definition of Civic Issue Graphs, followed by our detailed approach, consisting of scene graph generation and adversarial domain adaptation.
\\
%In order to generate a $CG$ from a scene graph model, we first require the model to appropriately understand and ground the objects ($O^1 \cup O^2$) and their corresponding predicates  ($p_{i \rightarrow j}$) which are relevant to the civic domain. We propose to use Adversarial Discriminative Domain Adaption (ADDA) to adapt existing scene graph model for our new task. We propose a Discriminator that learns to distinguish between the set of relations ($R_{CG}$) contained in $CG$ and the set of relations on which the model was trained ($R$) and use it to generalize the performance of the model for our task. Using additional training, we further refine the model to increase the probability of generating a predicate label for all object pairs $(o_1,o_2)$ where $o_1 \in O_1, o_2 \in O_2$ as defined above. Finally, we implement a Decision-Tree based classifier to construct the $CG$ using the relations returned by the model. Given an image $I$, the classifier identifies a final set of relations, $R^I_{CG}$) from $R_{CG}$ which are relevant for the image and uses them to construct the final $CG$

\noindent \textbf{Formal Definition:}
A scene graph is a structured representation of objects and the relationships between them present in an image. It consists of triples or relations (used inter-changeably) of the form $[object1, predicate, object2]$ where $predicate$ defines the relationship between the two objects and both $object1$ and $object2$ are grounded to their respective bounding box representations in the image. 
While a scene graph provides a complete representation of the contents of the scene in an image, our proposed Civic Issue Graph ($CG$) only consists of objects conveying a civic issue, their bounding boxes, and the predicate between these objects. We use the following notations to define a $CG$:
% It consists of a set of bounding boxes $B$, object set $O$ and a set of binary relations $R$ \cite{neuralmotif}.

\begin{itemize}
\item $B = \{b_1,\dots,b_n\}$: Set of bounding boxes $b_i \in \mathbb{R}^4$; $b_i$ represents the bounding box for an object $i$, defined as $b_i = (x, y, w, h)$, where $x$ and $y$ are co-ordinates of the centre of the bounding box, and $w$ and $h$ are the width and height of the bounding box.
\item $O^1 = \{o^1_1,\dots,o^1_n\}$: Set of objects essential for defining a civic issue, 
%{\color{red} and in CG these objects acts as a '\textit{subject}' or '\textit{head}'}
, \textit{e.g.}, `\textit{pothole}', `\textit{garbage}', \textit{etc}.
\item $O^2 = \{o^2_1,\dots,o^2_n\}$: Set of objects that define the context of objects in $O^1$, %{\color{red} and in CG these objects acts as a '\textit{object}' or '\textit{tail}'} 
\textit{e.g.}, `\textit{street}', `\textit{building}', \textit{etc}.
\item $O = O^1 \cup O^2$: Set of all objects that assign a class label $o_i \in O$ to each $b_i$
\item $P = \{p_1,\dots,p_n\}$: Set of predicates defining geometric or position-based relationships between $o^1_i \in O^1$ and $o^2_i \in O^2$, \textit{e.g.}, `\textit{above}', `\textit{next\_to}', `\textit{in}', \textit{etc}.
\item $R_{CG} = \{r_1,\dots,r_n\}$: Set of $CG$ relations with nodes $(b_i,o^1_i) \in B \times O^1$, $(b_j,o^2_j) \in B \times O^2$, and predicate label $p_{i \rightarrow j} \in P$, \textit{e.g.}, $[garbage,on,street]$, where $\textit{`garbage'} \in O^1$, $\textit{`street'} \in O^2$, and $\textit{`on'} \in P$
\end{itemize}

\begin{figure}
  \includegraphics[width=0.47\textwidth, height = 1.2in]{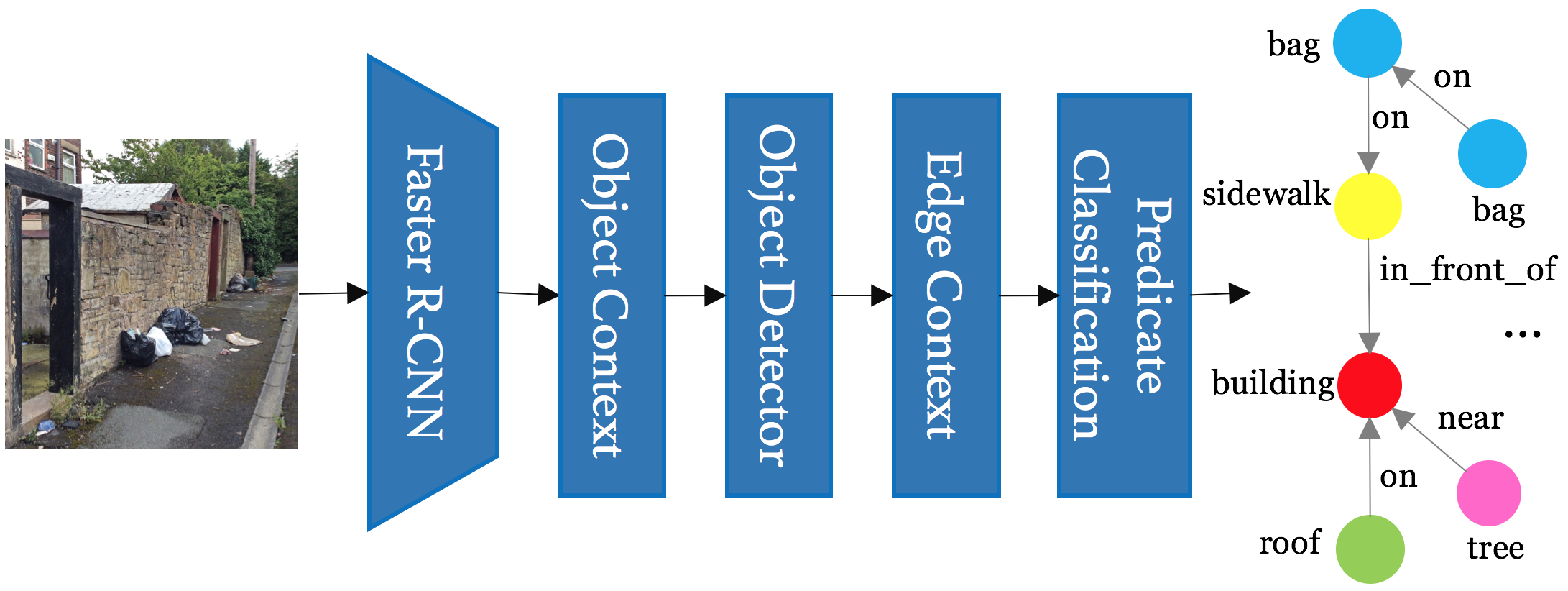}
  \caption{An overview of the MotifNet model}
  \label{fig:motif-net}
\end{figure}

\subsection{Scene Graph Generation}
% Scene graphs provide a complete structured representation of all objects in an image along with the relations between them. 
Several methods have been proposed for generating scene graphs from images and all of them require labelled training data \cite{messagepassing,neuralmotif}. The MotifNet model, proposed by Zellers et al. \cite{neuralmotif}, is the current state-of-the-art for generating scene graphs and we utilize this model for demonstrating our approach. However, our approach is generic and can be applied to other models with similar architecture as well.
\\

\noindent \textbf{MotifNet Model:}
As part of their approach, Zellers et al. highlight that the elements of a visual scene are often governed by the presence of high-level structural regularities, or motifs, such as, ``people tend to wear clothes''.
Such regularities indicate that given an image -- i) predicted object labels may depend on one another, and ii) predicted predicate labels may depend on the predicted object labels. Long Short-Term Memory (LSTM) networks \cite{hochreiter1997long} are known to capture such dependencies in the input sequence, when the gap between the dependencies is not known. The MotifNet model uses two bidirectional LSTMs \cite{zhang2015bidirectional} to -- i) capture the dependencies between object labels (referred as object context), and ii) capture the dependencies between the predicate labels and the object labels (referred as edge context). Fig \ref{fig:motif-net} presents a high-level overview of the model, which consists of:

\begin{figure*}
\centering
  \includegraphics[width=\textwidth, height = 2.3in]{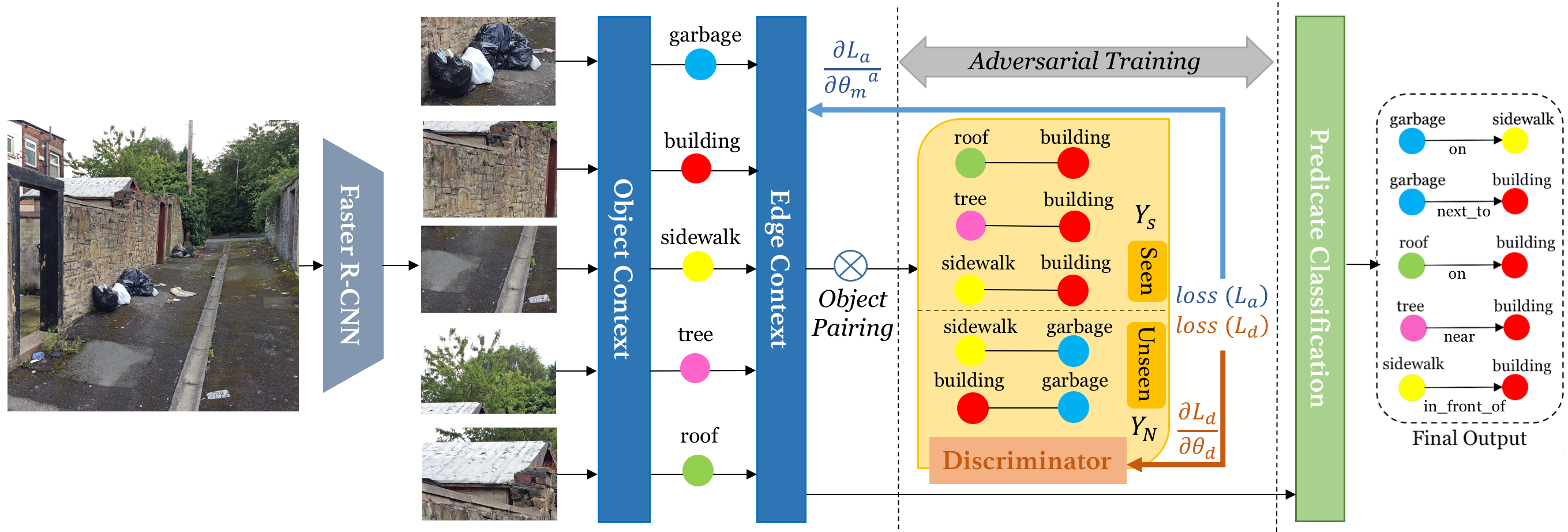}
  \caption{An illustration of our model: Faster R-CNN provides the object labels and their bounding regions. Object context generates a contextualized representation for each object. Edge context generates a contextualized representation for each edge using the representation of the object pairs (head and tail). During \textit{adversarial training}, information regarding the edge context is passed on to the Discriminator, which learns to distinguish between the $seen$ and $unseen$ object pairs. The training objective of the Discriminator results in gradients flowing into the Discriminator as well as the edge context layer of the MotifNet Model. The loss for the model decreases as the model learns to fool the Discriminator by adapting a uniform representation for $seen$ and $unseen$ classes.}
  \label{fig:model}
\end{figure*}

\begin{itemize}[leftmargin=*]
\item \textbf{Object Detection}: The MotifNet architecture consists of a Faster R-CNN model \cite{ren2015faster} to detect the objects present in an image. For each image $I$, the object detector provides a set of region proposals, $\mathbf{B} = [\mathbf{b}_1, \dots, \mathbf{b}_n]$. Each region proposal $\mathbf{b}_i$, is indicative of an object present in the image and is associated with a feature vector $\mathbf{f}_i$ and an object label probability vector $\mathbf{l}_i$.
\item \textbf{Object Context}: The MotifNet model uses bidirectional LSTM layers to construct a contextualized representation $\mathbf{C}$, for the set of region proposals $\mathbf{B}$. Here $\mathbf{C}$ models the dependencies between different region proposals. Eq. \ref{eq:objectcontext} shows the formulation of $\mathbf{C}$, in terms of $\mathbf{f}_i$, $\mathbf{l}_i$ and $\mathbf{W}_1$ where $\mathbf{W}_1$ is a parameter matrix that maps $\mathbf{l}_i$ to $\mathbb{R}^{100}$. 
% \todomj{**why R^100?**}

\item \textbf{Object Decoder}: The contextualized representation $\mathbf{C}$, is used to predict the final object labels $\mathbf{O}$. The labels ($\hat{\mathbf{o}}_i$) are decoded sequentially using another LSTM, where the hidden state for each label ($\mathbf{h}_i$) is conditioned on the previously decoded label (Eq. \ref{eq:lstm}). The hidden state is then used to compute the final object labels $\hat{\mathbf{o}}_{i}$ (Eq. \ref{eq:objectlabel}).
\item \textbf{Edge Context}: The model constructs another contextualized representation $\mathbf{D}$ using additional bidirectional LSTM layers, where $\mathbf{D}$ models the dependency between the relation labels and the object labels . Eq. \ref{eq:edgecontext} shows the formulation of $\mathbf{D}$, in terms of $\mathbf{c}_i$, $\hat{\mathbf{o}}_i$ and $\mathbf{W}_2$ where $\mathbf{W}_2$ is a parameter matrix that maps $\hat{\mathbf{o}}_{i}$ to $\mathbb{R}^{100}$.
\item \textbf{Predicate Classification}: For a sequence of region proposals ($\mathbf{B}$), quadratic number of object pairs are possible. An object pair $(b_i,b_j)$, is represented by the model using the final contextualized 
%{\color{red} head ($\mathbf{W}_h\mathbf{d}_i$) and tail ($\mathbf{W}_t\mathbf{d}_j$)} 
representations, ($\mathbf{d}_i,\mathbf{d}_j$) and the feature vector ($\mathbf{f}_{i,j}$) representing the union of these objects (Eq. \ref{eq:objectpairs}). Here $\mathbf{W}_h, \mathbf{W}_t$ project %{\color{red} context} 
$\mathbf{d}_i,\mathbf{d}_j$ into 
%{\color{red} \textit{head} and \textit{tail} space respectively of dimension} 
$\mathbb{R}^{4096}$. The model uses a softmax layer with this representation as input to identify the predicate label ($p_{i \rightarrow j}$) for each object pair or label it as background (Eq. \ref{eq:predicateclassification}). Here, $\mathbf{W}_t, \mathbf{w}_{o_i,o_j}$ represent the weights of the softmax layer. Object pairs with a valid predicate label (non-background) denote the final relations present in the scene graphs.
 \end{itemize}

Here is the mathematical formulation of the model:
% Hence probability of the scene graph will be:
% \begin{equation}
% \mathbf{P}(G|I) = \mathbf{P}(B|I)\:\mathbf{P}(O|B,I)\:\mathbf{P}(P|B,O,I)
% \end{equation}

%\subsubsection{Bounding Boxes}
%Faster R-CNN is used as a object detector which provides a set of region proposals $B$ for an image $I$. Each proposal contains a feature vector $\mathbf{f}_i$ and object label probability vector $\mathbf{l}_i$.
%\subsubsection{Object Context}
%Bidirectional LSTM are used to construct contextualized features on the set of proposal regions $B$.
\begin{equation} \label{eq:objectcontext}
\mathbf{C} = [\mathbf{c}_1, \dots, \mathbf{c}_n] = \mathbf{biLSTM({[\mathbf{f}_i;\mathbf{W}_1\mathbf{l}_i]}_{i=1,\dots,n})}
\end{equation}
% $\mathbf{C}=[\mathbf{c}_1, \dots, \mathbf{c}_n]$ denotes object context for every region in the set $B$.
%\subsubsection{Object Decoder}
%Object labels are sequentially decoded using Object Context $\mathbf{C}$ for each proposal region.
\begin{equation} \label{eq:lstm}
\mathbf{h}_i = \mathbf{LSTM}_i([\mathbf{c}_i; \hat{\mathbf{o}}_{i-1}])
\end{equation}
\begin{equation} \label{eq:objectlabel}
\hat{\mathbf{o}}_{i} = argmax(\mathbf{W}_o\mathbf{h}_{i})\:\:|\textrm{one hot vector}|
\end{equation}
% where $\hat{\mathbf{o}}_{i}$ is the one hot vector.
%\subsubsection{Edge Context}
%Contextualized representation for proposal regions $B$ and objects $O$  is constructed.
\begin{equation} \label{eq:edgecontext}
\mathbf{D} = [\mathbf{d}_1, \dots, \mathbf{d}_n] = \mathbf{biLSTM}_i({[\mathbf{c}_i; \mathbf{W}_2\hat{\mathbf{o}}_{i}]}_{i=1,\dots,n})
\end{equation}
% where $\mathbf{W}_2$ can be seen as word vectors.
%\subsubsection{Relations}
%For each possible  pair, i.e quadratic number of pairs, say $b_i$ and $b_j$, the probability of an edge $p_{i \rightarrow j}$ can be computed by using  edge context $\mathbf{d}_i$, $\mathbf{d}_j$ and a feature vector of the union of both boxes $\mathbf{f}_{i,j}$
\begin{equation} \label{eq:objectpairs}
\mathbf{g}_{i,j} = (\mathbf{W}_h\mathbf{d}_i)\circ (\mathbf{W}_t\mathbf{d}_j)\circ\mathbf{f}_{i,j}
\end{equation}
\begin{equation} \label{eq:predicateclassification}
\mathbf{P}(p_{i \rightarrow j}|B,O) = softmax(\mathbf{W}_r\mathbf{g}_{i,j} + \mathbf{w}_{o_i,o_j})
\end{equation}
% where $\mathbf{W}_h$ and $\mathbf{W}_t$ map head and tail context and $\mathbf{w}_{o_i, o_j}$ is bias specific for head and tail.

\subsection{Adversarial Domain Adaptation}
Domain adaption involves using an existing model trained on ``source'' domain where labelled data is available, and generalizing it to a ``target'' domain, where labelled data is not available. Domain adaptation has been helpful for tasks such as image captioning \cite{chen2017show} that require a large corpora of images and their labels, as getting this data for each and every domain is unfeasible. More recently, adversarial methods for domain adaptation \cite{da6} have also been proposed, where the training procedure is similar to the training of Generative Adversarial Networks (GANs)\cite{gan}. We present an adversarial training approach for a scene graph model, which, to the best of our knowledge, has not been explored before. Domain adaptation for scene graphs is challenging due to the large domain shift in the images as well as the feature space of relations (Fig. \ref{fig:domain-shift}). For instance, the Visual Genome dataset (VG) \cite{visualgenome} used for training scene graph models, consists of a mix of indoor and outdoor scenes with more object instances, whereas our dataset of civic issues consists of specific outdoor scenes depicting a civic issue. Moreover, some of the relations observed in the civic issue domain are not even present in the visual genome dataset (e.g., garbage-on-street). In the following subsections, we provide more details about our cross-domain setting followed by our approach for adversarial domain adaption.

\subsubsection{Cross-Domain Setting}
Scene graph models trained on a particular dataset can detect only those relations that are already $seen$ by the model, or in other words, present in the training dataset. For our task of generating $CG$, the model needs to detect ${R}_{CG}$, \textit{i.e.}, the set of relations contained in $CG$. Note that the set of relations in ${R}_{CG}$ can be further divided into ${R}_{s}$ and ${R}_{n}$, where ${R}_{s}$ is the set of relations previously $seen$ by the model, \textit{e.g.}: $[tree,over,fence]$ and ${R}_{n}$ is the set of relations previously $unseen$ by the model \textit{e.g.}: $[garbage,on,street]$. In the absence of any labelled data for ${R}_{n}$, we want to generalize the model already trained on ${R}_{s}$, to adapt to ${R}_{n}$ as well.

\subsubsection{Adversarial Approach}
Adversarial approach for domain adaptation consists of two models -- a pre-trained generator model and a discriminator model. In our setting, we use the MotifNet model pre-trained on VG dataset as the generator and propose a discriminator model that can distinguish between ${R}_{s}$ and ${R}_{n}$. During pre-training, the MotifNet model learns a representation for the object pairs (Eq. \ref{eq:objectpairs} and \ref{eq:predicateclassification}) which is used to predict the final set of relations (${R}_{s}$). Without adversarial training, the model has not learned the representation for any $unseen$ pair of objects from the civic domain and will not be able to predict such relations (${R}_{n}$).
Therefore, during adversarial training, the objective of the MotifNet model is to learn a mapping of target object pairs ($unseen$) to the feature space of the source object pairs ($seen$). This objective is supported via the discriminator, which is a binary classifier between the source and target domains. The MotifNet model can be said to have learned a uniform representation of object pairs corresponding to ${R}_{s}$ and ${R}_{n}$, if the classifier trained using this representation can no longer distinguish between ${R}_{s}$ and ${R}_{n}$. Therefore, we introduce two constrained objectives which seek to -- i) find the best discriminator model that can accurately classify ${R}_{s}$ and ${R}_{n}$, and ii) ``maximally confuse'' the discriminator model by learning new mapping for ${R}_{n}$. Once the source and target feature spaces are regularized, the predicate classifier trained on the $seen$ object pairs can be directly applied to $unseen$ object pairs, thereby eliminating the need for labelled training data.

Fig~\ref{fig:model} summarizes our adversarial training procedure. We first pre-train the MotifNet model on the VG dataset using cross-entropy loss and then update it using adversarial training. During adversarial training, the parameters for the MotifNet model and the discriminator are optimised according to a constrained adversarial objective. To optimize the discriminator model, we use the standard classification loss ($L_d$). In order to optimize the MotifNet model, we use the standard loss function ($L_a$) with inverted labels (seen $\rightarrow$ unseen, unseen $\rightarrow$ seen) thereby satisfying the adversarial objective. 
This entire training process is similar to the setting of GANs. 
We iteratively update the MotifNet model and the Discriminator with a ratio of $N_m$:$N_d$ with $N_m<N_d$, \textit{i.e.}, the Discriminator is updated more often than the MotifNet model. We now provide a mathematical formulation of our training approach.

\noindent \textbf{Discriminator}
We define the Discriminator as a binary classifier with $seen$ and $unseen$ as the two set of classes. For each object pair $(o_i,o_j)$, the Discriminator is provided with two inputs: 1) $\mathbf{g}_{i,j}$: final representation of the object pair generated by the model  and 2) $(\mathbf{W}_h\mathbf{d}_i)\circ (\mathbf{W}_t\mathbf{d}_j)$: contextualized representation of the object pair without the visual features. We further experimented with different inputs to the discriminator (details in Appendix).
%$(\mathbf{W}_h\mathbf{d}_i)\circ (\mathbf{W}_t\mathbf{d}_j)$ and $\mathbf{g}_{i,j}$.
%In the normal setting, the first representation is used to identify the predicate for each object pair.
The Discriminator consists of 2 fully connected layers, followed by a softmax layer to generate probability $C_d(l|o_i,o_j)$, where $l \in \{seen,unseen\}$.
The mathematical formulation of the discriminator for a given object pair ($o_i$, $o_j$) is:
\begin{equation}
\mathbf{F}_{i, j} = \mathbf{Dis([\mathbf{g}_{i,j};(\mathbf{W}_h\mathbf{d}_i)\circ (\mathbf{W}_t\mathbf{d}_j)])}
\end{equation}
\begin{equation}
\mathbf{C}_d = softmax(\mathbf{W}_d\mathbf{F}_{i, j} + \mathbf{b}_d)
\end{equation}

% where $l \in \{$problematic pair, non-problematic pair$\}$.
%Discriminator is trained as a supervised binary classification model.
\noindent \textbf{Training Discriminator}
% Loss for the Discriminator (D's objective)
Let ${OP}_{cv}$ be the set of all object pairs identified by the model for an image belonging to the civic domain $\mathbf{I_{cv}}$.
% \[{OP}_{vg} = NeuralMotif(\mathbf{I}_{vg})\]
\[{OP}_{cv} = MotifNet(\mathbf{I_{cv}})\]

\noindent The goal of the Discriminator is formulated as a supervised classification training objective:
% $\mathbf{I}_{vg}$ denotes image from Visual Genome Dataset,
% \begin{equation}
% % \mathcal{L}_d(\theta_d) =-\smashoperator{\sum_{{OP}_{cv},\:n=1}^{N_{cv}}}\log\mathbf{C}_d(l_{cv}^n|y^n) -\smashoperator{\sum_{{OP}_{vg},\:n=1}^{N_{vg}}}\log\mathbf{C}_d(l_{vg}^n|y^n)
% \mathcal{L}_d(\theta_d)=-\smashoperator{\sum_{{OP}_{cv},\:n=1}^{N}}\log\mathbf{C}_d(l^n|y^n)
% \end{equation}

\begin{equation}
% \mathcal{L}_d(\theta_d) =-\smashoperator{\sum_{{OP}_{cv},\:n=1}^{N_{cv}}}\log\mathbf{C}_d(l_{cv}^n|y^n) -\smashoperator{\sum_{{OP}_{vg},\:n=1}^{N_{vg}}}\log\mathbf{C}_d(l_{vg}^n|y^n)
\mathcal{L}_d(\theta_d)=-\smashoperator{\sum_{{OP}_{cv},\:n=1}^{N}}\log\mathbf{C}_d(l^n|y^n)
\end{equation}

\[  l^n = \left\{\begin{array}{ll}
    1 (seen) & \textrm{if }\ y^n \in Y_{s} \\[\jot]
    0 (unseen) & \textrm{if }\ y^n \in Y_{n},
  \end{array}\right.
\]
% \[l_{vg}^n = 1 \:\: :\: \forall\:\: n \:\in \:N_{vg} \]
where $y^n = (o_i, o_j)^n$, and $Y_n$ and $Y_{s}$ are the set of object pairs corresponding to $R_n$ and $R_s$, respectively. $\theta_{d}$ denotes the parameters of the Discriminator to be learned. We minimize $\mathcal{L}_d$ while training the discriminator.

\noindent \textbf{Training Model}
 In accordance with the inverted label loss described above, the training objective of the model is defined as follows:
%\[{OP}_{cv} = NeuralMotif(\mathbf{I}_{cv})\]
\begin{equation*}
\begin{aligned}
\mathcal{L}_a(\theta_m^a) =-\smashoperator{\sum_{{OP}_{cv},\:n=1}}^{N}\log\mathbf{C}_d(l^n|y^n) & & \forall y^n \in Y_{n}
\end{aligned}
\end{equation*}
% \[  l^n = \left\{\begin{array}{ll}
%     1 (unseen) & \textrm{if }\ y^n \in Y_{n},
%   \end{array}\right.
% \]
Here $\theta_m^a: \{\mathbf{W}_h, \mathbf{W}_t\}$
denotes the parameters of the model that are updated during adversarial training. We minimize $\mathcal{L}_a$ while updating the model.

\section{Experiments and Evaluation}
The simplest approach to identify the civic issue from images is to classify them into a predefined set of categories. We first report the performance of the baseline classifier which categorizes input images into different civic issue categories. The results show the limitations of a classification-based approach for handling images depicting a wide range of civic issues. 
Following this, we provide the implementation details of our model. We define a set of metrics which are derived from the standard metrics used for scene graph evaluation, for appropriately evaluating our approach. We conduct multiple experiments and provide generic insights for adversarial training of scene graph models. Finally, using human evaluation, we establish the efficacy of our model in appropriately representing civic issues from images.
\\
%Third, we conduct an ablation study to evaluate our model at different experimental settings and also establish the contribution of different components in our approach.
%Finally, we evaluate the output of our model with the help of human annotations and establish the efficacy of our model to appropriately represent the civic issue.

\noindent \textbf{Classification Approach}
% \subsubsection{Setup}
We trained a classifier (using VGG-16 network pre-trained on MS COCO dataset) to categorize images using the set of ten most frequent categories as defined on FixMyStreet complaint forum \cite{fixmystreet}. The classifier was trained on 80640 images and tested using 4992 images. 

\begin{table}[ht!]
\centering
\begin{tabular}{@{} c *1c @{}}
\toprule
 \multicolumn{1}{l}{Category}    &  Test Accuracy\\ 
\midrule \midrule
Potholes & 82.59 \\
Fly-tipping & 81.34 \\
Street/Traffic light &  68.37 \\
Graffiti & 64.73 \\
Pavements & 52.0 \\
Road traffic signs & 31.89 \\
Roads &  \textbf{16.42} \\
Garbage &  \textbf{15.59}  \\
Drainage/Manhole  & \textbf{7.84}  \\
Street Cleaning & \textbf{4.04} \\
\bottomrule
\end{tabular}
\caption{Class-wise Accuracy for the classifier}
\label{table:cat}
\end{table}

On the test data, this model achieves an accuracy of 47.13\%\footnote{Please refer to Appendix for more details on the classifier accuracy}, with F1-score of 38.76. Table \ref{table:cat} shows the class-wise accuracy for the classifier. While the accuracy for the three most accurate classes were 86.5\%, 83.6\% and 75.2\%, 4 out of 10 classes had their accuracy less than 17\%. Such large variation in the accuracy for different classes indicates that classifying images into different categories is not sufficient. 

%By applying transfer learning on VGG-16, we trained a classifier on 80640 training images to predict the issue category for a given image. 

%\textit{E.g.}, there is confusion between \textit{Garbage} and \textit{Fly-tipping} categories. Both categories represent different issues
%and between \textit{Roads} and \textit{Potholes}. In both these pairs, the two categories are very similar, and there is no clear distinction in terms of the civic issue they define. Even images corresponding to such overlapping categories tend to have very similar visual features. This shows that classifying civic issues into such pre-defined categories can be ambiguous, which makes this approach unsuitable for appropriately understanding civic issues from images.

\subsection{Implementation Details}
\subsubsection{Data Preprocessing}
For all our experiments, we use the datasets defined in Section \ref{sec:dataset}. In order to train the Discriminator, it requires a set of examples corresponding to the two classes: $R_n$ and $R_s$. Using the dataset-2, we extracted the set of relations ($R_{CG}$) and considered the 150 most frequent triples from this list. We manually refined this set by removing erroneous triples (\textit{e.g.:} $[pothole,on,building]$) and adding new triples based on existing triples (\textit{e.g.:} $[garbage,on,sidewalk] \Rightarrow [sidewalk,has,garbage]$). This resulted in 130 triples which are classified as follows: the triples for which the object pair is previously seen by the model, \textit{i.e.}, it is present in the VG dataset, are classified as  $R_{s}$ (80 out of 130 triples), and the rest 50 triples are classified as $R_{n}$.
For predicate fine-tuning, we use the same set of 130 triples. 
From dataset-2, we use 90\% (7384) of the images for updating the model, and the remaining 820 as test set, which is used for reporting experimental results and for the final human evaluation. 

\begin{figure}
\centering
\includegraphics[width=0.45\textwidth, height =1.6in]{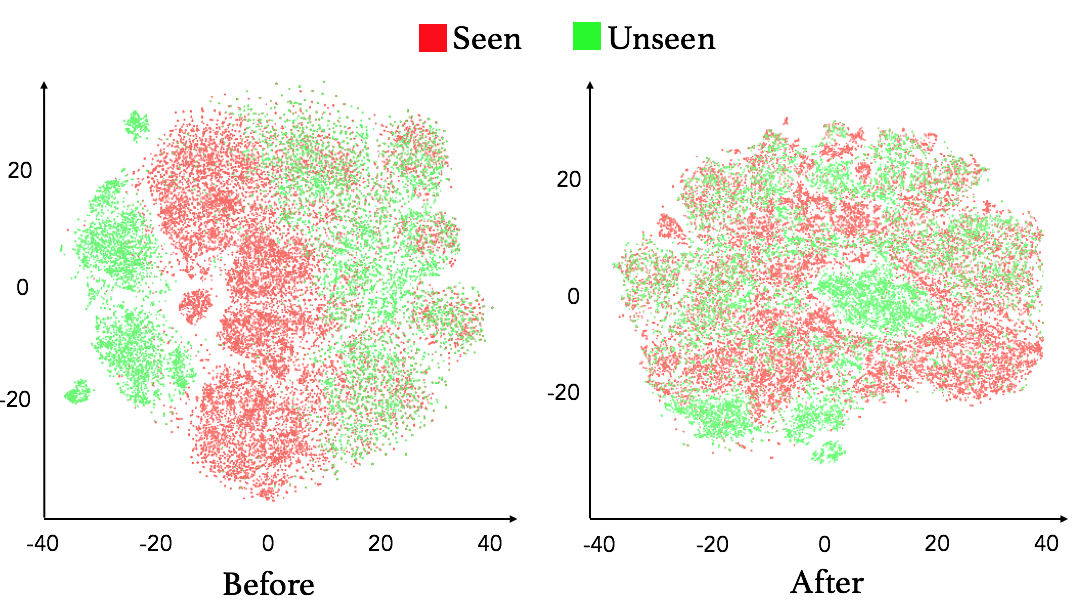}
\label{fig:f2}
\caption{Comparison between the representations (t-SNE embeddings) for $seen$ and $unseen$ object pairs before and after Adversarial Adaptation}
  \label{fig:domain-shift}
\end{figure}

\subsubsection{Faster R-CNN Training}
For the model to detect the objects in the civic domain, we train a Faster R-CNN model for the 19 object classes (present in the Dataset-2). 14 of these classes such as tree, building, street, \textit{etc}. are already present in VG dataset, and we utilize that for our training. For the remaining 5 classes, such as garbage, pothole, \textit{etc}., we use the dataset-1. The number of samples of a class from the VG dataset is much higher compared to the number of samples for a class in our new dataset. While training, we ensure an upper limit of 8000 and a lower limit of 3000 on the sample size for each class through a combination of under-sampling and over-sampling. The Faster R-CNN is trained for 10 epochs using SGD optimizer on 3 GPUs, with a batch size of 18 and a learning rate of 1.8 x $10^{-2}$, which was reduced to 1.8 x $10^{-3}$ after validation mAP plateaus.

% \subfloat[Visual Features]{\includegraphics[width=0.23\textwidth, height =2in]{figures/obj_pairs.png}\label{fig:f1}}
%   \hfill
%   \subfloat[Edge Context]{ 

\subsubsection{Scene Graph Model Pre-training}
We train the MotifNet model on a subset of VG dataset. We consider the 19 object classes (same as Faster R-CNN) and a (manually) filtered set of 32 predicate classes which are commonly found in the civic domain. We use the Faster R-CNN model trained on the civic domain for object detection. In the final setting, the model is trained without the `Object Decoder' and the difference is highlighted as part of experimental results. The rest of the training setup is same as the original MotifNet model (described in \cite{neuralmotif}), with the model being trained for 32 epochs. Please see the Appendix more details on pre-training of the MotifNet model.
%12 epochs for scene graph classification and 20 epochs refining the scene graph detection. 
%We also trained the model by removing the decoder layer from the model, using the same object labels predicted by faster R-CNN. Object decoder was helpful on Visual Genome dataset as body parts like nose, head, finger are not well detected by faster r-cnn but using object context body parts can be detected. But in our dataset, these objects are not needed to be detected.

\begin{figure*}
\centering
  \includegraphics[width=\textwidth, height = 2.2in]{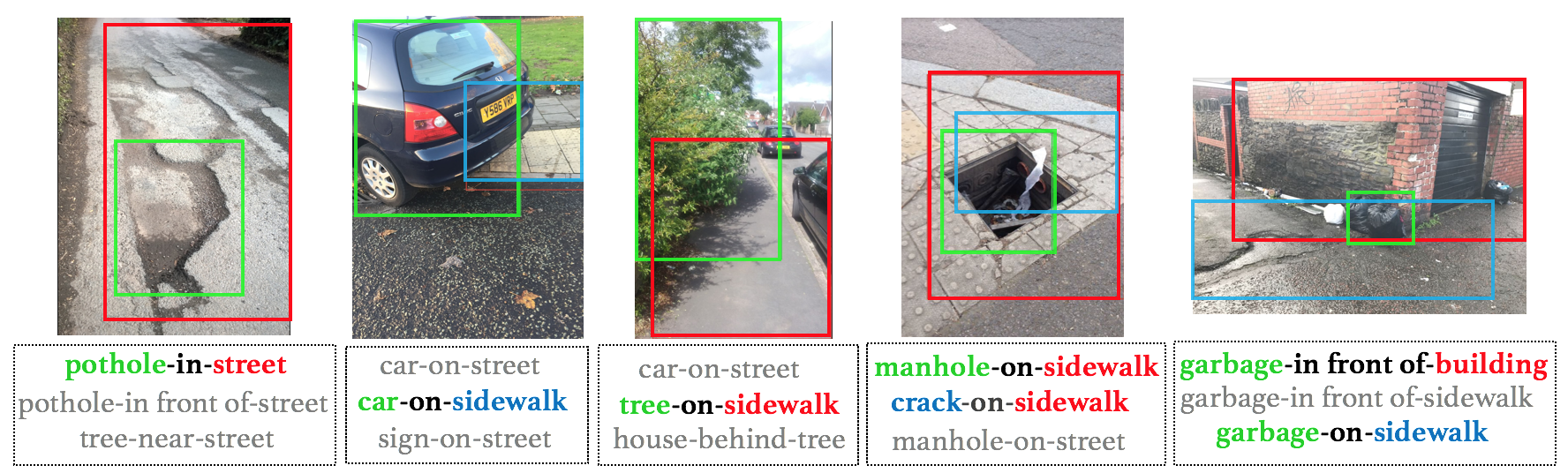}
  \caption{Qualitative examples presenting the Civic Issue Graphs generated by our model. We show the top 3 relations and highlight the ones that are representative of the civic issue along with their bounding regions}
  \label{fog:examples}
\end{figure*}

\subsubsection{Adversarial Training}
Discriminator used in adversarial training consists of 3 fully connected layers: two layers with 4096 hidden units followed by the final softmax output. Each hidden layer is followed by a batch normalization, leakyReLU activation function with negative slope of 0.2 and apply a dropout in the training phase with keeping probability of 0.5. 
%We use the pre-trained MotifNet model with ${\theta_m}^a = \{\mathbf{W}_h, \mathbf{W}_t\}$ denoting parameters of the model which are updated during adversarial training. %We further experiment with two different settings of ${\theta_m}^a$ and compare their performance as part of the ablation study. 
Both discriminator and model are trained using ADAM optimizer with a learning rate of 1.2 x $10^{-2}$ and 1.2 x $10^{-3}$, respectively. The value of $N_d$ is set to 150 steps, while $N_m$ is set to 50 steps, with the model and the discriminator being trained iteratively for 12 epochs.

\subsection{Evaluation Metrics}
Previous work \cite{messagepassing} defines three different modes for analyzing a scene graph model: Predicate Classification ({\scshape PredCls}), Scene Graph Classification ({\scshape SGCls}), and Scene Graph Generation ({\scshape SGgen}). {\scshape PredCls} task examines the performance of the model for detecting the predicate, given a set of object pairs, in isolation from other factors. {\scshape SGCls} task measures the performance of the model for predicting the right object labels and predicates, given a set of localized objects. In {\scshape SGgen} task, the model has to simultaneously detect the set of objects and predict the right predicate for each pair of objects. 
For our approach of generating Civic Issue Graph ($CG$) using existing scene graph models, it is appropriate to report: (i) the performance of the existing model when generalized to this new domain, and (ii) the accuracy of the output $CG$ for representing the civic issue in the image.
Deriving from the existing set of tasks, we define a new set of tasks which can help in evaluating our model along these dimensions: 
\begin{itemize}
\item \textbf{\textsc{OPCls}}: the task is to predict the set of object pairs which are indicative of the civic issue present in the image.
\item \textbf{\scshape{CGCls}}: the task is to predict the set of relations which can represent the civic issue present in the image.
\item \textbf{{\scshape CGGen}}: the task is to simultaneously detect the region in the image and predict the right relations which are indicative of the civic issue.
\end{itemize}
For task \textbf{{\scshape OPCls}}, we report the experimental results, and use human evaluation for the task \textbf{{\scshape CGCls}} and \textbf{{\scshape CGGen}}.
In accordance with previous work, for \textbf{{\scshape OPCls}}, we report results for the image-wise recall metrics (R@k). Since our task is to predict object pairs which are found in civic domains, we report results for $R@1$, $R@5$, $R@10$ \& $R@20$ metrics. For \textbf{{\scshape CGCls}} and \textbf{{\scshape CGGen}}, we report the results using both Precision and Recall metrics ($k:\{1,3,5\}$)
%reported as part of the human evaluation to establish if the output $CG$ can appropriately represent the civic issue. %As stated earlier, the purpose of the $CG$ is to provide a representation of the civic issue instead of providing a complete description of the scene. Therefore, we for all the tasks, we report image wise precision metric $P@k$, where $P@k$ denotes the fraction of top $k$ predictions output by the model which are representative of the civic issue.

\subsection{Experimental Results}
%We evaluate our model at different experimental settings and present the findings. 
%We first present an evaluation of the model at different experimental settings and analyze the results. Next we use the model with the best performance and highlight the impact of each of our component: adversarial training \& predicate fine-tuning. 
%The results are reported for the task \textbf{} on the validation set. 

\subsubsection{Removing Object Decoder}
The MotifNet model after adversarial training performed poorly when tested on the images from civic domain ($R@10=10.5$). We found that the object decoder is not able to predict the correct object labels when the input image contains new objects from the civic domain, as the model has not been trained on these labels. On removing the decoder during test time (denoted as $MotifNet_{Adv}$ in the table), the performance improves significantly ($R@10=76.0$, Table \ref{table:precision}). Adapting the decoder to a new domain requires ground-truth data in terms of the sequence of objects and the labels, which may not be possible for the civic domain. Therefore, we decided to pre-train the MotifNet model without the decoder (denoted by $MotifNet^{wd}$) and directly use the object labels predicted by the Faster R-CNN. On updating the new model using adversarial training (denoted by $MotifNet_{Adv}^{wd}$), the performance improved significantly, particularly for $R@1$ and $R@5$. Table \ref{table:precision} shows the comparison between the different settings with $MotifNet_{Adv}^{wd}$ performing significantly better than all other models, for all the metrics.

\subsubsection{Adversarial Training vs Fine-tuning}
Results from the previous experiment shows that using adversarial training can significantly improve the performance, as the model has now been generalized to both $seen$ and $unseen$ classes. As an alternative approach, we also try to adapt the pre-trained model to our new domain by fine-tuning the predicate classification in the model. Mathematically, we aim to increase the value of $\mathbf{P}(p_{i \rightarrow j}|B,O)$), where $(o_i,p_{i \rightarrow j},o_j)$ correspond to $R_{CG}$. The training objective for this phase is defined as: 
\begin{equation}
\mathcal{L}_f(\theta_m^f) =-\smashoperator{\sum_{{OP}_{cv},\:n=1}}^{N}l\log\mathbf{P}(p_{i \rightarrow j}^n|B^n,O^n)
\end{equation}
\[  l = \left\{\begin{array}{ll}
    1 & \textrm{if }\ p_{i \rightarrow j}^n \in R_{CG} \\[\jot]
    0 & \textrm{otherwise},
  \end{array}\right.
\]
where $\theta_m^f: \{\mathbf{W}_r, \mathbf{w}_{o_i,o_j}\}$, \textit{i.e.}, the weights and bias of the predicate classifier of the model. We minimize $L_f$ while fine-tuning the model which is trained for 6 epochs. Table \ref{table:precision} shows that fine-tuning a pre-trained MotifNet model ($MotifNet_{fine}^{wd}$) brings slight improvement in the performance when compared to the original model ($MotifNet$). However, the model with adversarial training ($MotifNet_{Adv}^{wd}$) performs significantly better than the fine-tuned model ($MotifNet_{fine}^{wd}$).  Fine-tuning the model will only improve the detection of relations which are already $seen$ by the model, while adversarial training will generalize the performance across both $seen$ and $unseen$ classes. This is further highlighted in Fig. \ref{fig:domain-shift} that shows how the difference between the representations of $seen$ and $unseen$ relations has reduced through adversarial training. Further fine-tuning the adversarially updated MotifNet ($MotifNet_{Adv+fine}^{{wd}}$) model brings no improvement in the performance.  
\begin{table}[ht!]
\centering
\begin{tabular}{@{} l *4c @{}}
\toprule
 \multicolumn{1}{l}{Model Settings}    & R@1  & R@5  & R@10  & R@20 \\ 
\midrule \midrule
\textsc{$MotifNet$} & 35.6 & 64.9  & 75.4 & 79.7   \\
\textsc{$MotifNet_{Adv}$} & 37.7  & 65.7 & 76.0 & 79.8 \\
% \midrule
\textsc{$MotifNet^{wd}$}  & 37.7  & 63.0  & 73.3  & 78.9 \\
% \begin{tabular}[c]{@{}c@{}}\textsc{$\mathbf{{MN^{wd}}_{Adv}}$}\\ \end{tabular}
\textbf{\textsc{$MotifNet^{wd}_{Adv}$}} & \textbf{43.3} & \textbf{67.7} & \textbf{76.3} & \textbf{80.2} \\\midrule
\textsc{$MotifNet_{fine}^{wd}$}  & 38.9 & 63.6 & 73.8 & 79.2 \\
\textsc{$MotifNet_{Adv+fine}^{wd}$} &  43.1 & 67.7 & 76.3 & 80.2 \\
% \begin{tabular}[c]{@{}c@{}}MN without decoder\\ with Adv (EC)\end{tabular} & 18.7   & 47   & 59.3   & 70 \\
% \begin{tabular}[c]{@{}c@{}}MN without decoder\\ with Adv (only $\mathbf{g}_{i,j} $)\end{tabular} & 38   & 63.1  & 73.3   & 79.2 \\\midrule
% \begin{tabular}[c]{@{}c@{}}MN without decoder\\  with PredFine \end{tabular}    & \textbf{38.9} & \textbf{63.6} & \textbf{73.8} & \textbf{79.2} \\
\bottomrule
\end{tabular}
\caption{Recall for different settings; ${Adv}$: Adversarial Training; $fine$: fine-tuning; ${wd}$: \textit{without decoder} setting}
\label{table:precision}
\end{table}

\subsection{Human Evaluation}
To establish the efficacy of our model at appropriately representing civic issues from images, we asked Amazon Mechanical Turk workers to evaluate the output of our model. We randomly sampled 300 images from the test set; each image was evaluated by 3 workers. In accordance with our definition of $CG$, we filtered the final set of relations generated by our model and kept only the top 5 relations for which $o_i \in O_1$ and $o_j \in O_2$ where $(o_i,o_j)$ denotes the unordered set of objects in a relation (refer Section \ref{formaldef}). 
%For each image, we retrieved only relations with objects $o_i$ \& $o_j$ which adhere to the definition of $CG$ -- $o_i \in O_1$ and $o_j \in O_2$ (as defined in Section \ref{formaldef}). We further removed all duplicate relations. We retrieved the top 5 relations after filtering. 

The evaluation was carried out for the two tasks -- \textbf{\textsc{CGCls}} and \textbf{\textsc{CGGen}} in two phases. For the task \textbf{\textsc{CGCls}}, workers were shown an image along with 5 relations and were asked to select 0 or more relations that appropriately represent the civic issue(s) in that image along with an option to specify any additional relations separately. For the task \textbf{\textsc{CGGen}}, we retrieved the set of relations which were marked as relevant for a given image. For each such relation, the workers were shown the bounding regions for the objects present in the relation, and asked to evaluate the coverage of these bounding regions, on a scale of 0 to 10. We report two metrics -- Precision and Recall, for both the tasks and consider only the majority voted relations with a minimum average rating of 5 for the bounding regions. Table \ref{table:results} shows the performance of our model. The results show that 83.3\% of the times, the relation representing a civic issue is present in the top 3 relations of our $CG$, and 53.0\% of the times, the top relation itself represents a civic issue in the image. The accuracy on the \textbf{\textsc{CGGen}} task further indicates that our model is capable of generating accurate groundings for the objects representing the civic issue.

\begin{table}[ht!]
\centering
\begin{tabular}{@{} c *6c @{}}
\toprule
   & \multicolumn{3}{c}{\textbf{{\scshape CGCls}}} & \multicolumn{3}{c}{\textbf{{\scshape CGGen}}} \\ 
   \cmidrule[\heavyrulewidth](r){2-4} \cmidrule[\heavyrulewidth](r){5-7}
   & @1      & @3      & @5     & @1      & @3      & @5     \\  \toprule
Precision & 53.0     & 31.9     & 24.7    & 50.9     & 30.3     & 24.1    \\
Recall & 53.0     & 84.0     & 99.0    & 50.9     & 83.3     & 99.0   \\
\bottomrule
\end{tabular}
\caption{Precision and Recall values for the tasks \textbf{{\scshape CGCls}} and \textbf{{\scshape CGGen}} based on human evaluation}
\label{table:results}
\end{table}

\section{Discussion}
\label{discussion}
The approach we presented in this paper can be utilized in existing platforms, which allow users to report civic issues using images. Once the user uploads an image, our model can automatically generate text-based relations (\textit{e.g.,} garbage-on-street, garbage-next to-building) depicting the civic issue in the input image. These text-based relational descriptions can be shared with the authorities, which can be utilized for large scale analysis, thereby automating the process and removing any dependency on the actual image uploaded by the user. Furthermore, if needed, natural language descriptions can be generated from these relations using a template-based approach. If the confidence of our model is low, the user can be asked to verify the output generated by the model, before sending it to the authorities. Data collected in this process can be further used for retraining the model to improve its performance.
\newline \noindent \textbf{Limitations:} 
While our model can understand a wide range of civic issues from images, some issue categories either cannot be captured using images or require additional information in text to adequately report the issue.
For example, irregular water supply problem, car speeding on the road, \textit{etc.} requires text to report the number of days of irregular water supply or the car number plate details.

The importance of scene graph representation of images has already been proven for several tasks, including semantic image retrieval \cite{johnson2015image} and visual question answering \cite{visualgenome}. However, previous approaches rely on extensive ground truth annotations to train the scene graph model. This has limited the scope of scene graphs in domains where obtaining such annotated data is either unfeasible or costly. For instance, in the education domain, the semantic understanding of an image through a scene graph representation (\textit{e.g.}, bat-has-wings, bat-inside-cave) can support learning through automatic generation of picture stories and image-based assessments. In the fashion domain, scene graphs can be used to create ontologies with objects such as accessories, clothes, and more. Even though we present a specific application of generating Civic Issue Graphs, our presented approach and the insights gained from our experiments can help expand the generation of scene graphs for other domains as well, by reducing the dependency on extensive ground truth annotations.

% In such cases, reporting the issue and/or providing additional details using text can be helpful for both citizens as well as the authorities.

\section{Conclusion}
\label{conclusion}
We introduce a novel unsupervised mechanism of adapting existing scene graph models via adversarial training and present an application of our approach for generating Civic Issue Graph. The Civic Issue Graph can provide a complete representation for images with different types of civic issues, to help bridge the gap between images and text descriptions used to report issues. Our experimental analysis helps provide a framework for adapting scene graph models to other settings as well. We also release two multi-modal (text and images) datasets with information of civic issues, to encourage future work in this domain.

\appendix
\section{Appendix}
% \subsection{Faster R-CNN}
% % \subsection{Dataset}
% % \subsection{Training}
% Faster R-CNN gets a mAP score of 40.42 (at 50\% IoU) on a combined dataset which consists of our dataset (Table \ref{table:dataset}) and a subset of Visual Genome dataset (Table \ref{})
%50% IoU) on Visual Genome;
% \begin{table}[ht!]
% \centering
% \begin{tabular}{@{} c *4c @{}}
% \toprule
% \multicolumn{1}{c}{} & IoU  & Area & Max Dets  & Score \\ 
% \midrule\midrule
% Average Precision  & 0.50:0.95 & all & 100  & 0.260 \\ 
% Average Precision  & 0.50 & all & 100  & 0.404\\ 
% Average Precision  & 0.75 & all & 100  & 0.261 \\ 
% Average Precision  & 0.50:0.95 & medium & 100  & 0.186 \\ 
% Average Precision  & 0.50:0.95 & large & 100  & 0.265 \\ \midrule
% Average Recall  & 0.50:0.95 & all & 1  & 0.313 \\ 
% Average Recall  & 0.50:0.95 & all & 10  & 0.499 \\ 
% Average Recall  & 0.50:0.95 & all & 100  & 0.500 \\ 
% Average Recall  & 0.50:0.95 & medium & 100  & 0.379 \\ 
% Average Recall  & 0.50:0.95 & large & 100  & 0.526 \\ 
% \bottomrule
% \end{tabular}
% \caption{Faster R-CNN scores}
% \label{table:fr-cnn}
% \end{table}
\subsection{MotifNet Model}
% \subsection{Dataset}
We use 32 predicates classes and 19 object classes to train the MotifNet model on a subset of Visual Genome dataset. Table ~\ref{table:obj} and ~\ref{table:pred} show the frequencies of objects and predicates used for the training of MotifNet model.
\begin{table}[ht!]
\centering
\begin{tabular}{@{} c *3c @{}}
\toprule
\multicolumn{1}{c}{Object} & Frequency & Object & Frequency\\ 
\midrule\midrule
animal  &  3611 & bag  & 7391 \\ 
bottle  &  6246  & box &  5467\\
building  &  31805 & car  &  17352 \\ 
fence  &  12027 & house  &   5006\\ 
letter  &  6630 & pole  &  21205 \\ 
sidewalk  & 9478 & sign  &  23499  \\ 
street  &  10996 & tree &  49902 \\
crack  &  1313  & garbage  &   217\\ 
pothole  & 19  & manhole  &   179\\ 
\bottomrule
\end{tabular}
\caption{Object Frequency for training MotifNet}
\label{table:obj}
\end{table}
\begin{table}[ht!]
\centering
\begin{tabular}{@{} c *3c @{}}
\toprule
\multicolumn{1}{c}{Predicate} & Frequency & Predicate & Frequency\\ 
\midrule\midrule
above  &   47341 & across  &   1996\\ 
against  &   3092 & along  &   3624\\ 
at  &   9903 & attached to  &   10190\\ 
behind  &   41356 & between  &   3411\\ 
carrying  &   5213 & covered in  &   2312\\ 
covering  &   3806 & flying in  &   1973\\ 
from  &   2945 & growing on  &   1853\\ 
hanging from  &   9894 & has  &   277936\\ 
in  &   251756 & in front of  &   13715\\ 
laying on  &   3739 & lying on  &   1869\\ 
mounted on  &   2253 & near  &   96589\\ 
on  &   712409 & on back of  &   1914\\ 
over  &   9317 & painted on &   3095\\ 
parked on  &   2721 & part of  &   2065\\ 
sitting on  &   18643 & standing on &   14185\\ 
under  &   22596 & with &   66425\\ 
\bottomrule
\end{tabular}
\caption{Predicate Frequency for training MotifNet}
\label{table:pred}
\end{table}

% \subsection{Training}
We trained 2 variants of MotifNet model: \textit{with object decoder} and \textit{without object decoder}. Table ~\ref{table:motif} shows the evaluation of both the models on the VG dataset. 
\begin{table}[ht!]
\centering
\begin{tabular}{@{} c *3c @{}}
\toprule
\multicolumn{1}{c}{Model} & R@20 & R@50 & R@100\\ 
\midrule \midrule
$MotifNet-VG$  &  24.56 & 28.08  & 30.18 \\ 
$MotifNet^{wd}-VG$ &  22.49  & 26.58 &  28..35\\
\bottomrule
\end{tabular}
\caption{MotifNet model Results on Visual Genome dataset}
\label{table:motif}
\end{table}

\subsection{Classification}
Table \ref{table:classification} highlights the confusion between the different classes of the classifier. The confusion occurs mostly because the images describing different types of civic issues can have very similar visual features (\textit{e.g.} Roads and Potholes) and some categories of issues only differ in their semantic interpretation (\textit{e.g.} Garbage and Fly-tipping). Despite the similarities, it is still important to maintain this distinction in order to understand the nature of civic issues. Different categories often have different resolution process and may involve different authorities. 
   \begin{table}[ht!]
    \centering
    \begin{tabularx}{\columnwidth}{|p{2.8cm}|p{1.4cm}|p{1.4cm}|p{1.4cm}|}
        \hline
        \multirow{1}{*}{True Class} &
           \multicolumn{3}{|c|}{Predicted Class (\%)} \\
        \hline
        $C_1$: Potholes & $C_1$, 86.5 & $C_6$, 4.9 & $C_4$, 4.2  \\
        \hline
        $C_2$: Street light & $C_2$, 75.2 & $C_3$, 6.4 & $C_7$, 5.9  \\
        \hline
        $C_3$: Fly-tipping & $C_3$, 83.6 & $C_6$, 4.9 & $C_5$,  4.5  \\
        \hline
        $C_4$: Roads & $\mathbf{C_1}$, \textbf{39.7} & $C_4$, 20.7 & $\mathbf{C_6}$, \textbf{18.2} \\
        \hline
        $C_5$: Garbage & $\mathbf{C_3}$, \textbf{57.8} & $C_5$, 25.3 & $C_6$, 7.6 \\
        \hline
        $C_6$: Pavements & $C_6$, 53.6 & $C_1$, 19.5 & $C_3$, 14.2 \\
        \hline
%         Graffiti & Graffiti, 60.37 & Fly-tipping, 15.15 & Pavements, 7.17 \\
%         \hline
        $C_7$: Road signs & $C_7$, 37.2 & $C_2$, 17.6 & $C_3$, 6.0 \\
        \hline
        $C_8$: Drainage & $\mathbf{C_1}$, \textbf{30.3} & $\mathbf{C_6}$, \textbf{20.4} & $C_4$, 16.5 \\
        \hline
        $C_9$: Street cleaning & $\mathbf{C_3}$, \textbf{33.4} & $\mathbf{C_6}$, \textbf{23.5} & $C_1$, 15.5 \\
        \hline
      \end{tabularx}
%       \end{adjustbox}
    \caption{Top three predictions per class, representing the confusion matrix}
    \label{table:classification}
    \end{table}

\subsection{Changing the input of the Discriminator}
While updating the model using adversarial training (denoted by $MotifNet_{Adv}^{wd}$) the input to the discriminator is:  $[\mathbf{g}_{i,j};(\mathbf{W}_h\mathbf{d}_i)\circ (\mathbf{W}_t\mathbf{d}_j)]$ as mentioned in the paper. Here, $(\mathbf{W}_h\mathbf{d}_i)\circ (\mathbf{W}_t\mathbf{d}_j)$ denotes the contextualized representation of an object pair $\mathbf{(o_i,o_j)}$ generated by the model. 
We also tested the model using another input for the discriminator: $\mathbf{g}_{i,j}$, which is a dot product of the contextualized representation $((\mathbf{W}_h\mathbf{d}_i)\circ (\mathbf{W}_t\mathbf{d}_j))$ and the visual features $(f_{i,j})$ for the object pairs. However, in the second case, with $\mathbf{g}_{i,j}$ as the input (denoted by $MotifNet_{Adv}^{wd*}$, Table ~\ref{table:result}) the improvements in the score were much less.
    
% \item \textbf{Changing the weights of Motifnet}: We also kept weights of Bidirectional LSTM which is used to construct edge context in the model for updatation during adversarial training.

\begin{table}[ht!]
\centering
\begin{tabular}{@{} c *4c @{}}
\toprule
 \multicolumn{1}{l}{Model Settings}    & R@1  & R@5  & R@10  & R@20 \\ 
\midrule \midrule
\textsc{$MotifNet_{Adv}^{wd}$}  & {43.3} & {67.7} & 76.3 & 80.2 \\
% \textsc{$\mathbf{MN_{Adv, EC}}$} & 18.7   & 47   & 59.3   & 70 \\
\textsc{$MotifNet_{Adv}^{wd*}$} & 38   & 63.1  & 73.3   & 79.2 \\
\bottomrule
\end{tabular}
\caption{Recall for different inputs to the Discriminator; $\mathbf{Adv}$ denotes Adversarial Training, $*$ denotes the setting with $\mathbf{g}_{i,j}$ as input, and $wd$ denotes without decoder setting} \label{table:result}
\end{table}

\bibliographystyle{ACM-Reference-Format}
\bibliography{bibliography}


\begin{thebibliography}{00}

%%% the Web address.

\ifx \showCODEN    \undefined \def \showCODEN     #1{\unskip}     \fi
\ifx \showDOI      \undefined \def \showDOI       #1{{\tt DOI:}\penalty0{#1}\ }
  \fi
\ifx \showISBNx    \undefined \def \showISBNx     #1{\unskip}     \fi
\ifx \showISBNxiii \undefined \def \showISBNxiii  #1{\unskip}     \fi
\ifx \showISSN     \undefined \def \showISSN      #1{\unskip}     \fi
\ifx \showLCCN     \undefined \def \showLCCN      #1{\unskip}     \fi
\ifx \shownote     \undefined \def \shownote      #1{#1}          \fi
\ifx \showarticletitle \undefined \def \showarticletitle #1{#1}   \fi
\ifx \showURL      \undefined \def \showURL       #1{#1}          \fi
% invisible to TeX
\providecommand\bibfield[2]{#2}
\providecommand\bibinfo[2]{#2}
\providecommand\natexlab[1]{#1}

\bibitem[\protect\citeauthoryear{??}{fix}{2016}]%
        {fixmystreet}
 \bibinfo{year}{2016}\natexlab{}.
\newblock \showarticletitle{FixMyStreet}.
\newblock  (\bibinfo{year}{2016}).
\newblock
\showURL{%
\url{https://www.fixmystreet.com/}}


\bibitem[\protect\citeauthoryear{??}{may}{2017}]%
        {mayor-report}
 \bibinfo{year}{2017}\natexlab{}.
\newblock \bibinfo{title}{Mayor's Management Report}.
\newblock
  \url{https://www1.nyc.gov/assets/operations/downloads/pdf/mmr2017/2017_mmr.pdf}.
    (\bibinfo{year}{2017}).
\newblock


\bibitem[\protect\citeauthoryear{Agostino}{Agostino}{2013}]%
        {agostino}
\bibfield{author}{\bibinfo{person}{Deborah Agostino}.}
  \bibinfo{year}{2013}\natexlab{}.
\newblock \showarticletitle{Using social media to engage citizens: A study of
  Italian municipalities}.
\newblock \bibinfo{journal}{{\em Public Relations Review\/}}
  \bibinfo{volume}{{39}, 3} (\bibinfo{year}{2013}), \bibinfo{pages}{232--234}.
\newblock


\bibitem[\protect\citeauthoryear{Atreja, Aggarwal, Mohapatra, Dumrewal, Basu,
  and Dasgupta}{Atreja et~al\mbox{.}}{2018}]%
        {citicafe}
\bibfield{author}{\bibinfo{person}{Shubham Atreja}, \bibinfo{person}{Pooja
  Aggarwal}, \bibinfo{person}{Prateeti Mohapatra}, \bibinfo{person}{Amol
  Dumrewal}, \bibinfo{person}{Anwesh Basu}, {and} \bibinfo{person}{Gargi~B
  Dasgupta}.} \bibinfo{year}{2018}\natexlab{}.
\newblock \showarticletitle{Citicafe: An Interactive Interface for Citizen
  Engagement}. In \bibinfo{booktitle}{{\em 23rd International Conference on
  Intelligent User Interfaces}}. ACM, \bibinfo{pages}{617--628}.
\newblock


\bibitem[\protect\citeauthoryear{Bousmalis, Trigeorgis, Silberman, Krishnan,
  and Erhan}{Bousmalis et~al\mbox{.}}{2016}]%
        {da4}
\bibfield{author}{\bibinfo{person}{Konstantinos Bousmalis},
  \bibinfo{person}{George Trigeorgis}, \bibinfo{person}{Nathan Silberman},
  \bibinfo{person}{Dilip Krishnan}, {and} \bibinfo{person}{Dumitru Erhan}.}
  \bibinfo{year}{2016}\natexlab{}.
\newblock \showarticletitle{Domain separation networks}. In
  \bibinfo{booktitle}{{\em Advances in Neural Information Processing Systems}}.
  \bibinfo{pages}{343--351}.
\newblock


\bibitem[\protect\citeauthoryear{Chen, Liao, Chuang, Hsu, Fu, and Sun}{Chen
  et~al\mbox{.}}{2017a}]%
        {da7}
\bibfield{author}{\bibinfo{person}{Tseng-Hung Chen}, \bibinfo{person}{Yuan-Hong
  Liao}, \bibinfo{person}{Ching-Yao Chuang}, \bibinfo{person}{Wan~Ting Hsu},
  \bibinfo{person}{Jianlong Fu}, {and} \bibinfo{person}{Min Sun}.}
  \bibinfo{year}{2017}\natexlab{a}.
\newblock \showarticletitle{Show, Adapt and Tell: Adversarial Training of
  Cross-Domain Image Captioner.}. In \bibinfo{booktitle}{{\em ICCV}}.
  \bibinfo{pages}{521--530}.
\newblock


\bibitem[\protect\citeauthoryear{Chen, Liao, Chuang, Hsu, Fu, and Sun}{Chen
  et~al\mbox{.}}{2017b}]%
        {chen2017show}
\bibfield{author}{\bibinfo{person}{Tseng-Hung Chen}, \bibinfo{person}{Yuan-Hong
  Liao}, \bibinfo{person}{Ching-Yao Chuang}, \bibinfo{person}{Wan~Ting Hsu},
  \bibinfo{person}{Jianlong Fu}, {and} \bibinfo{person}{Min Sun}.}
  \bibinfo{year}{2017}\natexlab{b}.
\newblock \showarticletitle{Show, Adapt and Tell: Adversarial Training of
  Cross-Domain Image Captioner.}. In \bibinfo{booktitle}{{\em ICCV}}.
  \bibinfo{pages}{521--530}.
\newblock


\bibitem[\protect\citeauthoryear{Chen, Li, Sakaridis, Dai, and Van~Gool}{Chen
  et~al\mbox{.}}{2018}]%
        {da15}
\bibfield{author}{\bibinfo{person}{Yuhua Chen}, \bibinfo{person}{Wen Li},
  \bibinfo{person}{Christos Sakaridis}, \bibinfo{person}{Dengxin Dai}, {and}
  \bibinfo{person}{Luc Van~Gool}.} \bibinfo{year}{2018}\natexlab{}.
\newblock \showarticletitle{Domain Adaptive Faster R-CNN for Object Detection
  in the Wild}. In \bibinfo{booktitle}{{\em Computer Vision and Pattern
  Recognition (CVPR)}}.
\newblock


\bibitem[\protect\citeauthoryear{Cocchia}{Cocchia}{2014}]%
        {cocchia2014smart}
\bibfield{author}{\bibinfo{person}{Annalisa Cocchia}.}
  \bibinfo{year}{2014}\natexlab{}.
\newblock \showarticletitle{Smart and digital city: A systematic literature
  review}.
\newblock In \bibinfo{booktitle}{{\em Smart city}}. Springer,
  \bibinfo{pages}{13--43}.
\newblock


\bibitem[\protect\citeauthoryear{Dahlgren}{Dahlgren}{2011}]%
        {dahlgren2011-modalities}
\bibfield{author}{\bibinfo{person}{Peter Dahlgren}.}
  \bibinfo{year}{2011}\natexlab{}.
\newblock \showarticletitle{Parameters of online participation: Conceptualising
  civic contingencies}.
\newblock \bibinfo{journal}{{\em Communication management quarterly\/}}
  \bibinfo{volume}{{21}, 4} (\bibinfo{year}{2011}), \bibinfo{pages}{87--110}.
\newblock


\bibitem[\protect\citeauthoryear{Eriksson, Girod, Hull, Newton, Madden, and
  Balakrishnan}{Eriksson et~al\mbox{.}}{2008}]%
        {pothole1}
\bibfield{author}{\bibinfo{person}{Jakob Eriksson}, \bibinfo{person}{Lewis
  Girod}, \bibinfo{person}{Bret Hull}, \bibinfo{person}{Ryan Newton},
  \bibinfo{person}{Samuel Madden}, {and} \bibinfo{person}{Hari Balakrishnan}.}
  \bibinfo{year}{2008}\natexlab{}.
\newblock \showarticletitle{The pothole patrol: using a mobile sensor network
  for road surface monitoring}. In \bibinfo{booktitle}{{\em Proceedings of the
  6th international conference on Mobile systems, applications, and services}}.
  ACM, \bibinfo{pages}{29--39}.
\newblock


\bibitem[\protect\citeauthoryear{Ganin and Lempitsky}{Ganin and
  Lempitsky}{2015}]%
        {da8}
\bibfield{author}{\bibinfo{person}{Yaroslav Ganin} {and}
  \bibinfo{person}{Victor Lempitsky}.} \bibinfo{year}{2015}\natexlab{}.
\newblock \showarticletitle{Unsupervised Domain Adaptation by Backpropagation}.
  In \bibinfo{booktitle}{{\em Proceedings of the 32Nd International Conference
  on International Conference on Machine Learning - Volume 37}}
  \bibinfo{series}{{\em (ICML'15)}}. JMLR.org, \bibinfo{pages}{1180--1189}.
\newblock
\showURL{%
\url{http://dl.acm.org/citation.cfm?id=3045118.3045244}}


\bibitem[\protect\citeauthoryear{Goodfellow, Pouget-Abadie, Mirza, Xu,
  Warde-Farley, Ozair, Courville, and Bengio}{Goodfellow et~al\mbox{.}}{2014}]%
        {gan}
\bibfield{author}{\bibinfo{person}{Ian Goodfellow}, \bibinfo{person}{Jean
  Pouget-Abadie}, \bibinfo{person}{Mehdi Mirza}, \bibinfo{person}{Bing Xu},
  \bibinfo{person}{David Warde-Farley}, \bibinfo{person}{Sherjil Ozair},
  \bibinfo{person}{Aaron Courville}, {and} \bibinfo{person}{Yoshua Bengio}.}
  \bibinfo{year}{2014}\natexlab{}.
\newblock \showarticletitle{Generative Adversarial Nets}.
\newblock In \bibinfo{booktitle}{{\em Advances in Neural Information Processing
  Systems 27}}, \bibfield{editor}{\bibinfo{person}{Z.~Ghahramani},
  \bibinfo{person}{M.~Welling}, \bibinfo{person}{C.~Cortes},
  \bibinfo{person}{N.~D. Lawrence}, {and} \bibinfo{person}{K.~Q. Weinberger}}
  (Eds.). Curran Associates, Inc., \bibinfo{pages}{2672--2680}.
\newblock
\showURL{%
\url{http://papers.nips.cc/paper/5423-generative-adversarial-nets.pdf}}


\bibitem[\protect\citeauthoryear{Hart and Staveland}{Hart and
  Staveland}{1988}]%
        {nasa-tlx}
\bibfield{author}{\bibinfo{person}{Sandra~G Hart} {and}
  \bibinfo{person}{Lowell~E Staveland}.} \bibinfo{year}{1988}\natexlab{}.
\newblock \showarticletitle{Development of NASA-TLX (Task Load Index): Results
  of empirical and theoretical research}.
\newblock \bibinfo{journal}{{\em Advances in psychology\/}}
  \bibinfo{volume}{52} (\bibinfo{year}{1988}), \bibinfo{pages}{139--183}.
\newblock


\bibitem[\protect\citeauthoryear{Hochreiter and Schmidhuber}{Hochreiter and
  Schmidhuber}{1997}]%
        {hochreiter1997long}
\bibfield{author}{\bibinfo{person}{Sepp Hochreiter} {and}
  \bibinfo{person}{J{\"u}rgen Schmidhuber}.} \bibinfo{year}{1997}\natexlab{}.
\newblock \showarticletitle{Long short-term memory}.
\newblock \bibinfo{journal}{{\em Neural computation\/}} \bibinfo{volume}{{9},
  8} (\bibinfo{year}{1997}), \bibinfo{pages}{1735--1780}.
\newblock


\bibitem[\protect\citeauthoryear{IChangeMyCity}{IChangeMyCity}{2012}]%
        {icmc-link}
\bibfield{author}{\bibinfo{person}{IChangeMyCity}.}
  \bibinfo{year}{2012}\natexlab{}.
\newblock \url{https://ichangemycity.com}.   (\bibinfo{year}{2012}).
\newblock


\bibitem[\protect\citeauthoryear{Johnson, Krishna, Stark, Li, Shamma,
  Bernstein, and Fei-Fei}{Johnson et~al\mbox{.}}{2015}]%
        {johnson2015image}
\bibfield{author}{\bibinfo{person}{Justin Johnson}, \bibinfo{person}{Ranjay
  Krishna}, \bibinfo{person}{Michael Stark}, \bibinfo{person}{Li-Jia Li},
  \bibinfo{person}{David Shamma}, \bibinfo{person}{Michael Bernstein}, {and}
  \bibinfo{person}{Li Fei-Fei}.} \bibinfo{year}{2015}\natexlab{}.
\newblock \showarticletitle{Image retrieval using scene graphs}. In
  \bibinfo{booktitle}{{\em Proceedings of the IEEE conference on computer
  vision and pattern recognition}}. \bibinfo{pages}{3668--3678}.
\newblock


\bibitem[\protect\citeauthoryear{Karakiza}{Karakiza}{2015}]%
        {karakiza}
\bibfield{author}{\bibinfo{person}{Maria Karakiza}.}
  \bibinfo{year}{2015}\natexlab{}.
\newblock \showarticletitle{The impact of social media in the public sector}.
\newblock \bibinfo{journal}{{\em Procedia-Social and Behavioral Sciences\/}}
  \bibinfo{volume}{175} (\bibinfo{year}{2015}), \bibinfo{pages}{384--392}.
\newblock


\bibitem[\protect\citeauthoryear{Klawonn and Heim}{Klawonn and Heim}{2018}]%
        {klawonn2018generating}
\bibfield{author}{\bibinfo{person}{Matthew Klawonn} {and} \bibinfo{person}{Eric
  Heim}.} \bibinfo{year}{2018}\natexlab{}.
\newblock \showarticletitle{Generating Triples with Adversarial Networks for
  Scene Graph Construction}.
\newblock \bibinfo{journal}{{\em arXiv preprint arXiv:1802.02598\/}}
  (\bibinfo{year}{2018}).
\newblock


\bibitem[\protect\citeauthoryear{Krishna, Zhu, Groth, Johnson, Hata, Kravitz,
  Chen, Kalantidis, Li, Shamma, et~al\mbox{.}}{Krishna et~al\mbox{.}}{2017}]%
        {visualgenome}
\bibfield{author}{\bibinfo{person}{Ranjay Krishna}, \bibinfo{person}{Yuke Zhu},
  \bibinfo{person}{Oliver Groth}, \bibinfo{person}{Justin Johnson},
  \bibinfo{person}{Kenji Hata}, \bibinfo{person}{Joshua Kravitz},
  \bibinfo{person}{Stephanie Chen}, \bibinfo{person}{Yannis Kalantidis},
  \bibinfo{person}{Li-Jia Li}, \bibinfo{person}{David~A Shamma}, {and}
  \bibinfo{person}{others}.} \bibinfo{year}{2017}\natexlab{}.
\newblock \showarticletitle{Visual genome: Connecting language and vision using
  crowdsourced dense image annotations}.
\newblock \bibinfo{journal}{{\em International Journal of Computer Vision\/}}
  \bibinfo{volume}{{123}, 1} (\bibinfo{year}{2017}), \bibinfo{pages}{32--73}.
\newblock


\bibitem[\protect\citeauthoryear{Li, Ouyang, Zhou, Wang, and Wang}{Li
  et~al\mbox{.}}{2017}]%
        {sg1}
\bibfield{author}{\bibinfo{person}{Yikang Li}, \bibinfo{person}{Wanli Ouyang},
  \bibinfo{person}{Bolei Zhou}, \bibinfo{person}{Kun Wang}, {and}
  \bibinfo{person}{Xiaogang Wang}.} \bibinfo{year}{2017}\natexlab{}.
\newblock \showarticletitle{Scene graph generation from objects, phrases and
  region captions}. In \bibinfo{booktitle}{{\em Proceedings of the IEEE
  Conference on Computer Vision and Pattern Recognition}}.
  \bibinfo{pages}{1261--1270}.
\newblock


\bibitem[\protect\citeauthoryear{Liang, Guo, Chang, and Chen}{Liang
  et~al\mbox{.}}{2018}]%
        {liang2018visual}
\bibfield{author}{\bibinfo{person}{Kongming Liang}, \bibinfo{person}{Yuhong
  Guo}, \bibinfo{person}{Hong Chang}, {and} \bibinfo{person}{Xilin Chen}.}
  \bibinfo{year}{2018}\natexlab{}.
\newblock \showarticletitle{Visual Relationship Detection with Deep Structural
  Ranking}.
\newblock  (\bibinfo{year}{2018}).
\newblock


\bibitem[\protect\citeauthoryear{Long, Cao, Wang, and Jordan}{Long
  et~al\mbox{.}}{2015}]%
        {da3}
\bibfield{author}{\bibinfo{person}{Mingsheng Long}, \bibinfo{person}{Yue Cao},
  \bibinfo{person}{Jianmin Wang}, {and} \bibinfo{person}{Michael~I Jordan}.}
  \bibinfo{year}{2015}\natexlab{}.
\newblock \showarticletitle{Learning transferable features with deep adaptation
  networks}.
\newblock \bibinfo{journal}{{\em arXiv preprint arXiv:1502.02791\/}}
  (\bibinfo{year}{2015}).
\newblock


\bibitem[\protect\citeauthoryear{Maeda, Sekimoto, Seto, Kashiyama, and
  Omata}{Maeda et~al\mbox{.}}{2018}]%
        {road1}
\bibfield{author}{\bibinfo{person}{Hiroya Maeda}, \bibinfo{person}{Yoshihide
  Sekimoto}, \bibinfo{person}{Toshikazu Seto}, \bibinfo{person}{Takehiro
  Kashiyama}, {and} \bibinfo{person}{Hiroshi Omata}.}
  \bibinfo{year}{2018}\natexlab{}.
\newblock \showarticletitle{Road Damage Detection Using Deep Neural Networks
  with Images Captured Through a Smartphone}.
\newblock \bibinfo{journal}{{\em arXiv preprint arXiv:1801.09454\/}}
  (\bibinfo{year}{2018}).
\newblock


\bibitem[\protect\citeauthoryear{Manning, Surdeanu, Bauer, Finkel, Bethard, and
  McClosky}{Manning et~al\mbox{.}}{2014}]%
        {corenlp}
\bibfield{author}{\bibinfo{person}{Christopher Manning}, \bibinfo{person}{Mihai
  Surdeanu}, \bibinfo{person}{John Bauer}, \bibinfo{person}{Jenny Finkel},
  \bibinfo{person}{Steven Bethard}, {and} \bibinfo{person}{David McClosky}.}
  \bibinfo{year}{2014}\natexlab{}.
\newblock \showarticletitle{The Stanford CoreNLP natural language processing
  toolkit}. In \bibinfo{booktitle}{{\em Proceedings of 52nd annual meeting of
  the association for computational linguistics: system demonstrations}}.
  \bibinfo{pages}{55--60}.
\newblock


\bibitem[\protect\citeauthoryear{Mearns, Simmonds, Richardson, Turner, Watson,
  and Missier}{Mearns et~al\mbox{.}}{2014}]%
        {socialmedia2}
\bibfield{author}{\bibinfo{person}{Graeme Mearns}, \bibinfo{person}{Rebecca
  Simmonds}, \bibinfo{person}{Ranald Richardson}, \bibinfo{person}{Mark
  Turner}, \bibinfo{person}{Paul Watson}, {and} \bibinfo{person}{Paolo
  Missier}.} \bibinfo{year}{2014}\natexlab{}.
\newblock \showarticletitle{Tweet my street: a cross-disciplinary collaboration
  for the analysis of local twitter data}.
\newblock \bibinfo{journal}{{\em Future Internet\/}} \bibinfo{volume}{{6}, 2}
  (\bibinfo{year}{2014}), \bibinfo{pages}{378--396}.
\newblock


\bibitem[\protect\citeauthoryear{Mergel}{Mergel}{2012}]%
        {seeclickfix}
\bibfield{author}{\bibinfo{person}{Ines Mergel}.}
  \bibinfo{year}{2012}\natexlab{}.
\newblock \showarticletitle{Distributed democracy: Seeclickfix. com for
  crowdsourced issue reporting}.
\newblock  (\bibinfo{year}{2012}).
\newblock


\bibitem[\protect\citeauthoryear{Mittal, Yagnik, Garg, and Krishnan}{Mittal
  et~al\mbox{.}}{2016b}]%
        {mittal2016spotgarbage}
\bibfield{author}{\bibinfo{person}{Gaurav Mittal}, \bibinfo{person}{Kaushal~B
  Yagnik}, \bibinfo{person}{Mohit Garg}, {and} \bibinfo{person}{Narayanan~C
  Krishnan}.} \bibinfo{year}{2016}\natexlab{b}.
\newblock \showarticletitle{Spotgarbage: smartphone app to detect garbage using
  deep learning}. In \bibinfo{booktitle}{{\em Proceedings of the 2016 ACM
  International Joint Conference on Pervasive and Ubiquitous Computing}}. ACM,
  \bibinfo{pages}{940--945}.
\newblock


\bibitem[\protect\citeauthoryear{Mittal, Agarwal, and Sureka}{Mittal
  et~al\mbox{.}}{2016a}]%
        {socialmedia1}
\bibfield{author}{\bibinfo{person}{Nitish Mittal}, \bibinfo{person}{Swati
  Agarwal}, {and} \bibinfo{person}{Ashish Sureka}.}
  \bibinfo{year}{2016}\natexlab{a}.
\newblock \showarticletitle{Got a Complaint?-Keep Calm and Tweet It!}. In
  \bibinfo{booktitle}{{\em International Conference on Advanced Data Mining and
  Applications}}. Springer, \bibinfo{pages}{619--635}.
\newblock


\bibitem[\protect\citeauthoryear{Nam and Pardo}{Nam and Pardo}{2011}]%
        {nam2011conceptualizing}
\bibfield{author}{\bibinfo{person}{Taewoo Nam} {and} \bibinfo{person}{Theresa~A
  Pardo}.} \bibinfo{year}{2011}\natexlab{}.
\newblock \showarticletitle{Conceptualizing smart city with dimensions of
  technology, people, and institutions}. In \bibinfo{booktitle}{{\em
  Proceedings of the 12th annual international digital government research
  conference: digital government innovation in challenging times}}. ACM,
  \bibinfo{pages}{282--291}.
\newblock


\bibitem[\protect\citeauthoryear{Neirotti, De~Marco, Cagliano, Mangano, and
  Scorrano}{Neirotti et~al\mbox{.}}{2014}]%
        {neirotti2014current}
\bibfield{author}{\bibinfo{person}{Paolo Neirotti}, \bibinfo{person}{Alberto
  De~Marco}, \bibinfo{person}{Anna~Corinna Cagliano}, \bibinfo{person}{Giulio
  Mangano}, {and} \bibinfo{person}{Francesco Scorrano}.}
  \bibinfo{year}{2014}\natexlab{}.
\newblock \showarticletitle{Current trends in Smart City initiatives: Some
  stylised facts}.
\newblock \bibinfo{journal}{{\em Cities\/}}  \bibinfo{volume}{38}
  (\bibinfo{year}{2014}), \bibinfo{pages}{25--36}.
\newblock


\bibitem[\protect\citeauthoryear{Oquab, Bottou, Laptev, and Sivic}{Oquab
  et~al\mbox{.}}{2014}]%
        {da1}
\bibfield{author}{\bibinfo{person}{Maxime Oquab}, \bibinfo{person}{Leon
  Bottou}, \bibinfo{person}{Ivan Laptev}, {and} \bibinfo{person}{Josef Sivic}.}
  \bibinfo{year}{2014}\natexlab{}.
\newblock \showarticletitle{Learning and transferring mid-level image
  representations using convolutional neural networks}. In
  \bibinfo{booktitle}{{\em Proceedings of the IEEE conference on computer
  vision and pattern recognition}}. \bibinfo{pages}{1717--1724}.
\newblock


\bibitem[\protect\citeauthoryear{Pei, Cao, Long, and Wang}{Pei
  et~al\mbox{.}}{2018}]%
        {da13}
\bibfield{author}{\bibinfo{person}{Zhongyi Pei}, \bibinfo{person}{Zhangjie
  Cao}, \bibinfo{person}{Mingsheng Long}, {and} \bibinfo{person}{Jianmin
  Wang}.} \bibinfo{year}{2018}\natexlab{}.
\newblock \showarticletitle{Multi-Adversarial Domain Adaptation}. In
  \bibinfo{booktitle}{{\em AAAI}}.
\newblock


\bibitem[\protect\citeauthoryear{Ren, He, Girshick, and Sun}{Ren
  et~al\mbox{.}}{2015}]%
        {ren2015faster}
\bibfield{author}{\bibinfo{person}{Shaoqing Ren}, \bibinfo{person}{Kaiming He},
  \bibinfo{person}{Ross Girshick}, {and} \bibinfo{person}{Jian Sun}.}
  \bibinfo{year}{2015}\natexlab{}.
\newblock \showarticletitle{Faster r-cnn: Towards real-time object detection
  with region proposal networks}. In \bibinfo{booktitle}{{\em Advances in
  neural information processing systems}}. \bibinfo{pages}{91--99}.
\newblock


\bibitem[\protect\citeauthoryear{Schuster, Krishna, Chang, Fei-Fei, and
  Manning}{Schuster et~al\mbox{.}}{2015}]%
        {schuster2015generating}
\bibfield{author}{\bibinfo{person}{Sebastian Schuster}, \bibinfo{person}{Ranjay
  Krishna}, \bibinfo{person}{Angel Chang}, \bibinfo{person}{Li Fei-Fei}, {and}
  \bibinfo{person}{Christopher~D Manning}.} \bibinfo{year}{2015}\natexlab{}.
\newblock \showarticletitle{Generating semantically precise scene graphs from
  textual descriptions for improved image retrieval}. In
  \bibinfo{booktitle}{{\em Proceedings of the fourth workshop on vision and
  language}}. \bibinfo{pages}{70--80}.
\newblock


\bibitem[\protect\citeauthoryear{Skoric, Zhu, Goh, and Pang}{Skoric
  et~al\mbox{.}}{2016}]%
        {skoric2016social}
\bibfield{author}{\bibinfo{person}{Marko~M Skoric}, \bibinfo{person}{Qinfeng
  Zhu}, \bibinfo{person}{Debbie Goh}, {and} \bibinfo{person}{Natalie Pang}.}
  \bibinfo{year}{2016}\natexlab{}.
\newblock \showarticletitle{Social media and citizen engagement: A
  meta-analytic review}.
\newblock \bibinfo{journal}{{\em New Media \& Society\/}}
  \bibinfo{volume}{{18}, 9} (\bibinfo{year}{2016}),
  \bibinfo{pages}{1817--1839}.
\newblock


\bibitem[\protect\citeauthoryear{Tzeng, Hoffman, Saenko, and Darrell}{Tzeng
  et~al\mbox{.}}{2017}]%
        {da6}
\bibfield{author}{\bibinfo{person}{Eric Tzeng}, \bibinfo{person}{Judy Hoffman},
  \bibinfo{person}{Kate Saenko}, {and} \bibinfo{person}{Trevor Darrell}.}
  \bibinfo{year}{2017}\natexlab{}.
\newblock \showarticletitle{Adversarial discriminative domain adaptation}. In
  \bibinfo{booktitle}{{\em Computer Vision and Pattern Recognition (CVPR)}},
  \bibinfo{volume}{Vol.~1}. \bibinfo{pages}{4}.
\newblock


\bibitem[\protect\citeauthoryear{Xu, Zhu, Choy, and Fei-Fei}{Xu
  et~al\mbox{.}}{2017}]%
        {messagepassing}
\bibfield{author}{\bibinfo{person}{Danfei Xu}, \bibinfo{person}{Yuke Zhu},
  \bibinfo{person}{Christopher~B Choy}, {and} \bibinfo{person}{Li Fei-Fei}.}
  \bibinfo{year}{2017}\natexlab{}.
\newblock \showarticletitle{Scene graph generation by iterative message
  passing}. In \bibinfo{booktitle}{{\em Proceedings of the IEEE Conference on
  Computer Vision and Pattern Recognition}}, \bibinfo{volume}{Vol.~2}.
\newblock


\bibitem[\protect\citeauthoryear{Yu and Salari}{Yu and Salari}{2011}]%
        {pothole2}
\bibfield{author}{\bibinfo{person}{X Yu} {and} \bibinfo{person}{E Salari}.}
  \bibinfo{year}{2011}\natexlab{}.
\newblock \showarticletitle{Pavement pothole detection and severity measurement
  using laser imaging}. In \bibinfo{booktitle}{{\em Electro/Information
  Technology (EIT), 2011 IEEE International Conference on}}. IEEE,
  \bibinfo{pages}{1--5}.
\newblock


\bibitem[\protect\citeauthoryear{Zellers, Yatskar, Thomson, and Choi}{Zellers
  et~al\mbox{.}}{2018}]%
        {neuralmotif}
\bibfield{author}{\bibinfo{person}{Rowan Zellers}, \bibinfo{person}{Mark
  Yatskar}, \bibinfo{person}{Sam Thomson}, {and} \bibinfo{person}{Yejin Choi}.}
  \bibinfo{year}{2018}\natexlab{}.
\newblock \showarticletitle{Neural Motifs: Scene Graph Parsing with Global
  Context}. In \bibinfo{booktitle}{{\em Proceedings of the IEEE Conference on
  Computer Vision and Pattern Recognition}}. \bibinfo{pages}{5831--5840}.
\newblock


\bibitem[\protect\citeauthoryear{Zhang, Zheng, Hu, and Yang}{Zhang
  et~al\mbox{.}}{2015}]%
        {zhang2015bidirectional}
\bibfield{author}{\bibinfo{person}{Shu Zhang}, \bibinfo{person}{Dequan Zheng},
  \bibinfo{person}{Xinchen Hu}, {and} \bibinfo{person}{Ming Yang}.}
  \bibinfo{year}{2015}\natexlab{}.
\newblock \showarticletitle{Bidirectional long short-term memory networks for
  relation classification}. In \bibinfo{booktitle}{{\em Proceedings of the 29th
  Pacific Asia Conference on Language, Information and Computation}}.
  \bibinfo{pages}{73--78}.
\newblock


\bibitem[\protect\citeauthoryear{Zhuang, Wu, Shen, Reid, and van~den
  Hengel}{Zhuang et~al\mbox{.}}{2018}]%
        {zhuang2018hcvrd}
\bibfield{author}{\bibinfo{person}{Bohan Zhuang}, \bibinfo{person}{Qi Wu},
  \bibinfo{person}{Chunhua Shen}, \bibinfo{person}{Ian~D Reid}, {and}
  \bibinfo{person}{Anton van~den Hengel}.} \bibinfo{year}{2018}\natexlab{}.
\newblock \showarticletitle{HCVRD: A Benchmark for Large-Scale Human-Centered
  Visual Relationship Detection.}. In \bibinfo{booktitle}{{\em AAAI}}.
\newblock


\end{thebibliography}

\end{document}